\documentclass[lettersize,journal]{IEEEtran}
\usepackage{amsmath,amsfonts}
\usepackage{algorithmic}
\usepackage{algorithm}
\usepackage{array}
\usepackage[caption=false,font=normalsize,labelfont=sf,textfont=sf]{subfig}
\usepackage{textcomp}
\usepackage{stfloats}
\usepackage{url}
\usepackage{verbatim}
\usepackage{graphicx}
\usepackage{cite}
\usepackage{color,xcolor}
\usepackage{balance}
\usepackage{textcomp}
\usepackage{stfloats}
\usepackage{verbatim}
\usepackage{hyperref}
\usepackage[misc]{ifsym}
\usepackage[numbers,sort&compress]{natbib}
\usepackage{amsmath,amsfonts}
\usepackage{multirow}
\usepackage{booktabs}
\usepackage{makecell}

\usepackage{tikz,xcolor,hyperref}
\definecolor{lime}{HTML}{A6CE39}
\DeclareRobustCommand{\orcidicon}{%
    \begin{tikzpicture}
    \draw[lime, fill=lime] (0,0)
    circle [radius=0.16]
    node[white] {{\fontfamily{qag}\selectfont \tiny ID}};    \draw[white, fill=white] (-0.0625,0.095)
    circle [radius=0.007];    \end{tikzpicture}
    \hspace{-2mm}}
\foreach \x in {A, ..., Z}{%
    \expandafter\xdef\csname orcid\x\endcsname{\noexpand\href{https://orcid.org/\csname orcidauthor\x\endcsname}{\noexpand\orcidicon}}
    }

\begin{document}

\title{PG-VTON: A Novel Image-Based Virtual Try-On Method via Progressive Inference Paradigm }

\newcommand{\orcidauthorA}{0000-0003-0145-1690}
\newcommand{\orcidauthorB}{0000-0001-9358-0099}
\newcommand{\orcidauthorD}{0000-0002-5003-3092}
\newcommand{\orcidauthorE}{0000-0002-8055-6468}

\author{Naiyu Fang\orcidA{}~\IEEEmembership{Graduate Student Member,~IEEE}, Lemiao Qiu\orcidB{}~\IEEEmembership{Member,~IEEE}, Shuyou Zhang, Zili Wang\orcidD{}~\IEEEmembership{Member,~IEEE}, Kerui Hu\orcidE{}
\thanks{This work has been submitted to the IEEE for possible publication. Copyright may be transferred without notice, after which this version may no longer be accessible.}
\thanks{This work was supported by the National Natural Science Foundation of China under Grant 52375271, the Natural Science Foundation of Zhejiang Province under Grant LY23E050011, and Pioneer and Leading Goose R\&D Program of Zhejiang under Grant 2022C01051. The Associate Editor coordinating the review of this manuscript and approving it for publication was Prof. Guoying Zhao. \it(Corresponding authors: Lemiao Qiu)}
\thanks{The authors are with State Key Laboratory of Fluid Power \& Mechatronic Systems, Zhejiang University, Hangzhou, 310027, China (e-mail: FangNaiyu@zju.edu.cn; qiulm@zju.edu.cn; zsy@zju.edu.cn; ziliwang@zju.edu.cn; hkr457@zju.edu.cn )}
}



\maketitle

\begin{abstract}
Virtual try-on is a promising computer vision topic with a high commercial value wherein a new garment is visually worn on a person with a photo-realistic effect. Previous studies conduct their shape and content inference at one stage, employing a single-scale warping mechanism and a relatively unsophisticated content inference mechanism. These approaches have led to suboptimal results in terms of garment warping and skin reservation under challenging try-on scenarios. To address these limitations, we propose a novel virtual try-on method via progressive inference paradigm (PGVTON) that leverages a top-down inference pipeline and a general garment try-on strategy. Specifically, we propose a robust try-on parsing inference method by disentangling semantic categories and introducing consistency. Exploiting the try-on parsing as the shape guidance, we implement the garment try-on via warping-mapping-composition. To facilitate adaptation to a wide range of try-on scenarios, we adopt a covering more and selecting one warping strategy and explicitly distinguish tasks based on alignment. Additionally, we regulate StyleGAN2 to implement re-naked skin inpainting, conditioned on the target skin shape and spatial-agnostic skin features. Experiments demonstrate that our method has state-of-the-art performance under two challenging scenarios. The code will be available at \url{https://github.com/NerdFNY/PGVTON}.
\end{abstract}


\section{Introduction}
\label{sec1}

\IEEEPARstart{V}{irtual} try-on is a promising topic for commercial applications in computer vision. The image-based virtual try-on has no requirements for professional 3d modeling of the person {\color{blue}\citep{ hu20213dbodynet,zhao20183}} and garment {\color{blue}\citep{sekhavat2016privacy}}. It still preset a photo-realistic wearing effect only conditioned on the person and garment image. However, as the variability of person and garment increases, the content inference and garment warping employed in previous studies have encountered challenges when faced with difficult try-on scenarios. In this paper, we propose a progressive inference paradigm of virtual try-on and employ advanced strategies of garment try-on and skin inpainting to enhance the try-on realism even in the presence of distinct garment categories or complex poses.

The objective of virtual try-on is to manipulate the shape of a new garment and wear it on a person, which involves several challenging techniques such as image warping, image inpainting, and image fusion. In the nascent stages of this field, Han X et al. and Wang B et al. provided two distinct pipelines to achieve this aim. VITON {\color{blue}\citep{ han2018viton}} proposed by Han X et al. first synthesized a coarse result conditioned on garment image and person representation and then warped the garment by TPS deformation to refine the result. CP-VTON {\color{blue}\citep{wang2018toward}} proposed by Wang B et al. firstly regressed TPS coefficients to obtain a coarse result and then synthesized a rendered person according to the coarse result and the person representations.

These two pipelines have a significant inspiration for subsequent studies. However, their try-on effects have not been satisfactory due to the unrealistic performance in the garment and body. The garment try-on task involves modifying the shape of a garment to the desired form while preserving its original style and category. Meanwhile, since the garment try-on affects the representation of the body, it is crucial that the inferred body maintains reasonable content and structure. To address this issue, numerous researchers have contributed to preserving garment texture and body features. Yu R et al. {\color{blue}\citep{yu2019vtnfp}} proposed a new image synthesis network to preserve garment and body part details. Yang H et al. {\color{blue}\citep{yang2020towards}} introduced second-order constraints into TPS deformation and implemented layout adaptation by a content fusion module. Minar M R et al. {\color{blue}\citep{minar2020cp}} proposed CP-VTON+ to adjust inputs and refine semantic categories based on CP-VTON. Neuberger A et al. {\color{blue}\citep{neuberger2020image}} proposed O-VITON and refined the appearance through online optimization. Ge C et al. {\color{blue}\citep{ge2021disentangled}} introduced cycle consistency into the self-supervised training and disentangled the garment and non-garment regions. He S et al. {\color{blue}\citep{he2022style}} and Bai S et al {\color{blue}\citep{bai2022single}} proposed Flow-Style and DAFlow to warp garments by estimating the flow.

The comprehensive research of virtual try-on emerges endlessly, such as changing try-on pose, trying on garments dressed on another person, and displaying try-on effects in other forms. To be specific, some studies combined pose transfer {\color{blue}\citep{fang2022novel}} with virtual try-on to wear a new garment and change the personal pose {\color{blue}\citep{ma2021fda, hu2022spg}}. Dong H et al. {\color{blue}\citep{dong2019towards}} proposed MGTON and realized the pose-guided virtual try-on. Xie Z et al. {\color{blue}\citep{xie2021towards}} proposed PASTA-GAN to implement patch-routed disentanglement in unsupervised training. Cui A et al. {\color{blue}\citep{cui2021dressing}} proposed DiOr to implement the tasks of virtual try-on, pose transfer, and fashion editing at the same time. Transferring the garment from one person to another person {\color{blue}\citep{raj2018swapnet, liu2019swapgan}} intend to enrich the garment form to alleviate the dependence on paired images. Raj A et al. {\color{blue}\citep{raj2018swapnet}} proposed SwapNet to first implement this task. Liu T et al. {\color{blue}\citep{liu2021spatial}} proposed SPATT to tackle the inferring unobserved appearance by establishing correspondence in UV space. Lewis K M et al. {\color{blue}\citep{lewis2021tryongan}} proposed TryOnGAN and improved the body shape deformation and skin color with a conditioned StyleGAN2. Some studies tried to alter the display form of the try-on effect to the high-resolution image, image sequence, and video. Dong H et al. {\color{blue}\citep{dong2019fw}} proposed FW-GAN to realize the video virtual try-on. Chen C Y et al. {\color{blue}\citep{chen2021fashionmirror}} proposed FashionMirror and synthesized try-on image sequences conditioned on the driven-pose sequence. Choi S et al. {\color{blue}\citep{choi2021viton}} proposed VITON-HD to increase the try-on resolution to 1024×768. In addition, some studies contribute to the end-to-end model for virtual try-on, which is conducive to direct inference without prior. Issenhuth T et al. {\color{blue}\citep{issenhuth2020not}} proposed WUTON to avoid human parsing in the student model. Ge Y et al. {\color{blue}\citep{ge2021parser}} further adjusted the inputs of the student model and distilled the appearance flows.

As previously discussed, achieving accurate garment try-on while simultaneously preserving body features is a key objective for virtual try-on research. Garment warping and content inference are the two main techniques utilized for achieving this objective. Three primary warping techniques are utilized for garment warping, namely STN {\color{blue}\citep{jaderberg2015advances}}, TPS deformation {\color{blue}\citep{bookstein1989principal}}, and optical flow {\color{blue}\citep{dosovitskiy2015flownet}}. STN learns the affine transformation to rigidly warp garment, while TPS deformation aligns control points with the energy function to non-rigidly warp garment in a low DOF. Optical flow estimates the pixel movement to implement high DOF warping. When the warped garment still not matches the target shape precisely or the skin re-exposes when trying a new category garment, it is imperative to infer the content of the garment and skin at the misalignment according. Previous studies {\color{blue}\citep{yang2020towards}} has relied on traditional CNN framework such as U-Net {\color{blue}\citep{ronneberger2015u}} for this task. However, vision transformer {\color{blue}\citep{dosovitskiy2020image}} (ViT) shows its superior in high-level computer vision topics as its self-attention mechanism. Since its quadratic computation overhead, it is a challenge to apply it to low-level topics like the aforementioned content inference task. Recent studies are breaking this deadlock. Swin transformer {\color{blue}\citep{liu2021swin}} calculated the attention with shifted window; Focal Transformer {\color{blue}\citep{yang2021focal}} captured the long-range and short-range dependence at coarse and fine-grain, respectively. Restomer {\color{blue}\citep{zamir2022restormer}} calculated the attention across the channel.

However, some issues are still impeding its commercial applications in previous studies. 1) no robust and explicit try-on parsing inference mechanism. Previous studies have attempted to infer the shape and content at one stage, without establishing an explicit try-on parsing inference mechanism, guiding to garment category alteration and arm blur at subsequent tasks; 2) unrealistic garment try-on under complex scenarios. It is attributed to a single-scale warping mechanism, the unadvanced mechanism and backbone in content inference, and the inexplicit joint of warping and inference, which have led to poor texture reservation and unrealistic try-on effect; 3) no specialized and well-designed skin inpainting mechanism. Previous studies treat garment inference and skin inpainting as a single task, which results in skin artifacts and skin color shifts.
To remedy these, we propose a progressive inference paradigm for robust and realistic virtual try-on under complex situations. ‘Progressive Inference Paradigm’ embodies two key aspects: 1) progressive pipeline. We first leverage the try-on parsing inference model (TPIM) to provide shape guidance for downstream tasks. Subsequently, we leverage the progressive try-on module (PTM) and re-naked skin inpainting module (RSIM) to handle garment try-on and skin inpainting; 2) progressive try-on. We implement garment try-on by the coarse warping, the fined mapping, and composition in turn, and explicitly distinguish these tasks with alignment. The contribution of our paper is three-fold:

(1) The progressive try-on for the general circumstance. We leverage a novel warping strategy that covers more and selects one for a better warping extent and texture reservation. And we leverage the consistency and well-designed supervision strategy to implement fined mapping by Restormer.

(2) StyleGAN2-based skin inpainting. We adapt StyleGAN2 to conduct the re-naked skin inpainting conditioned on the skin shape and spatial-agnostic skin features. And we introduce random erasure into the self-supervised training to simulate the actual case of arm exposure.

(3) Experiments demonstrate that our method has state-of-the-art performance under two challenging try-on scenarios compared to other baselines. PG-VTON has achieved 2.74 IS and 35.47 hyperIQA on the dataset {\color{blue}\citep{han2018viton, wang2018toward}}.

\section{Progressive virtual try-on}
\label{sec2}

\subsection{Outline}
\label{sec2.1}

\begin{figure}[!ht]
\centering
\includegraphics[scale=.23]{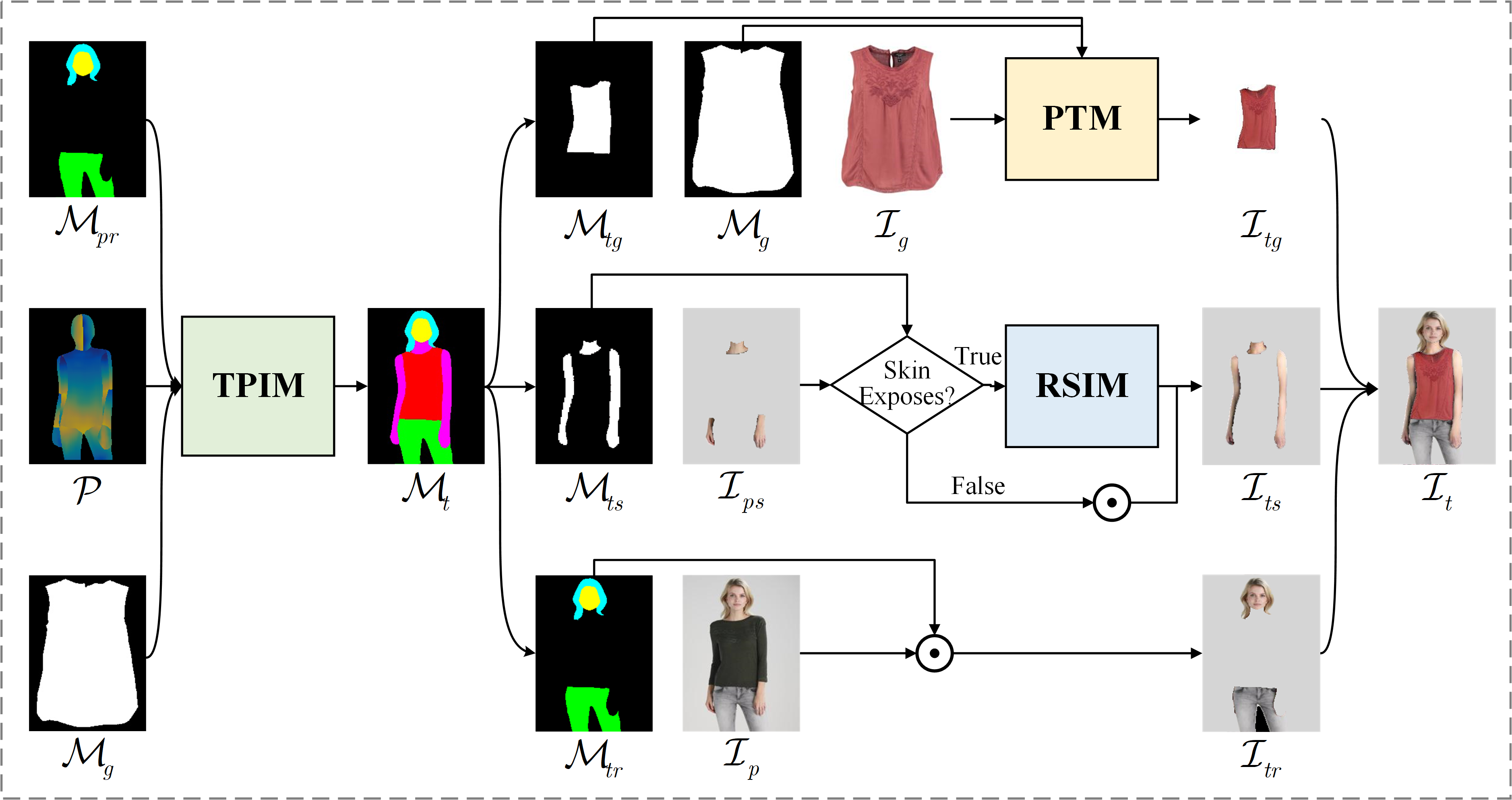}
\caption{The outline of PG-VTON. TPIM infers a try-on parsing for the downstream tasks. And then, PTM implements the garment try-on task by warping-mapping-composition while RSIM realizes the skin inpainting. Finally, we composite all components to yield a try-on result.}
\label{fig1}
\end{figure}

Virtual try-on intends to wear a new garment on an imaged person. To this end, as {\color{blue} Fig. \ref{fig1}} shows, we infer a try-on parsing ${{{\cal M}_t}}$ conditioned the new garment mask ${{{\cal M}_g}}$, the person parsing ${{{\cal M}_p}}$, and the person pose ${{\cal P}}$, and TPIM is responsible for this high-level task as ${{{\cal M}_t} = TPIM\left( {{{\cal M}_{pr}},{\cal P},{{\cal M}_g}} \right)}$, which is described in {\color{blue} Sec. \ref{sec2.2}}. Guided by ${{{\cal M}_t}}$, three low-level tasks follow behind. It is imperative to change this new garment to the target try-on shape. Therefore, the progressive try-on module leverages three stages to implement ${{{\cal I}_{tg}} = PTM\left( {{{\cal M}_{tg}},{{\cal M}_g},{{\cal I}_g}} \right)}$ by warping-mapping-composition, which is described in {\color{blue} Sec. \ref{sec2.3}}. The re-naked skin inferring is essential when new skin exposes in virtual try-on. Thus, RSIM implements ${{{\cal I}_{ts}} = RSIM\left( {{{\cal M}_{ts}},{{\cal I}_{ps}}} \right)}$ by adapting StyleGAN2 {\color{blue}\citep{ karras2020analyzing}}, which is described in {\color{blue} Sec. \ref{sec2.4}}. Furthermore, we multiply the remaining part of the person parsing ${{{\cal M}_{tr}}}$ with the person image ${{{\cal I}_p}}$ to directly obtain the remaining part of the try-on image ${{{\cal I}_{tr}}}$. Finally, we composite ${{{\cal I}_{tg}}}$, ${{{\cal I}_{ts}}}$, ${{{\cal I}_{tr}}}$ to form the try-on image ${{{\cal I}_t}}$.

As symbols vary widely, we depict naming rules of symbols for reading convenience. 1) Symbols ${{\cal M}}$, ${{\cal I}}$, ${{\cal P}}$ are the mask (parsing), the image, and the pose; 2) Subscripts ${g}$, ${g'}$, ${p}$, ${t}$ represent cases belonging to the new garment, the paired garment, the person, and the try-on result. The paired garment is the same as the dressed one on ${p}$ but is in-shop form; 3) Subscripts ${pg}$, ${ps}$, ${pr}$ are the dressed garment, the upper skin, and the remaining part subcategories of the person; 4) Subscripts ${tg}$, ${ts}$, ${tr}$ are same circumstances but in the try-on results; 5) The superscript ${ \sim }$ is the predicted result in the training phase, where the ground truth is the same symbol without ${ \sim }$.

Some significant signs and their descriptions are enumerated in {\color{blue} Table \ref{table1}}.

\begin{table}[!ht]
\caption{Signs and descriptions.}
\label{table1}
\resizebox{\columnwidth}{!}{
\begin{tabular}{@{}llll@{}}
\toprule
\multicolumn{1}{c}{Sign}     & \multicolumn{1}{c}{Description}                                  & \multicolumn{1}{c}{Sign} & \multicolumn{1}{c}{Description}              \\ \midrule
\multicolumn{2}{c|}{\textbf{For Person}}                                                        & \multicolumn{2}{c}{\textbf{For Garment}}                                \\
${{{\cal I}_p}}$             & \multicolumn{1}{l|}{Person Image}                                & ${{{\cal I}_g}}$         & New Garment Image                            \\
${{{\cal M}_p}}$             & \multicolumn{1}{l|}{Person Parsing}                              & ${{{\cal M}_g}}$         & New Garment Mask                             \\
${{\cal P}}$                 & \multicolumn{1}{l|}{Pose}                                        & ${{{\cal I}_{g'}}}$      & Same as ${{{\cal I}_{pg}}}$ but In-Shop Form \\
${{{\cal   I}_{pg}}}$        & \multicolumn{1}{l|}{Dressed Garment in   ${{{\cal I}_p}}$}       & ${{{\cal M}_{g'}}}$      & Same as ${{{\cal M}_{pg}}}$ but In-Shop Form \\ \cmidrule(l){3-4}
${{{\cal   M}_{pg}}}$        & \multicolumn{1}{l|}{Dressed Garment in   ${{{\cal M}_p}}$}       & \multicolumn{2}{c}{\textbf{For Try-on Result}}                          \\
${{{\cal   I}_{ps}}}$        & \multicolumn{1}{l|}{Upper Skin in ${{{\cal I}_p}}$}              & ${{{\cal I}_t}}$         & Try-on Image                                 \\
${{{\cal   M}_{ps}}}$        & \multicolumn{1}{l|}{Upper Skin in ${{{\cal M}_p}}$}              & ${{{\cal M}_t}}$         & Try-on Parsing                               \\
${{{\cal   I}_{pr}}}$        & \multicolumn{1}{l|}{Remaining Part in ${{{\cal I}_p}}$}          & ${{{\cal I}_{tg}}}$      & Dressed Garment in   ${{{\cal I}_t}}$        \\
${{{\cal   M}_{pr}}}$        & \multicolumn{1}{l|}{Remaining Part in ${{{\cal M}_p}}$}          & ${{{\cal M}_{tg}}}$      & Dressed Garment in   ${{{\cal M}_t}}$        \\
${{\tilde {\cal   I}_{ps}}}$ & \multicolumn{1}{l|}{Predicted result   of ${{{\cal   I}_{ps}}}$} & ${{{\cal I}_{ts}}}$      & Upper Skin in ${{{\cal I}_t}}$               \\
${{\tilde {\cal   M}_p}}$    & \multicolumn{1}{l|}{Predicted result   of ${{{\cal M}_p}}$}      & ${{{\cal M}_{ts}}}$      & Upper Skin in ${{{\cal M}_t}}$               \\
                             & \multicolumn{1}{l|}{}                                            & ${{{\cal I}_{tr}}}$      & Remaining Part in ${{{\cal I}_t}}$           \\
                             & \multicolumn{1}{l|}{}                                            & ${{{\cal M}_{tr}}}$      & Remaining Part in ${{{\cal M}_t}}$           \\ \bottomrule
\end{tabular}
}
\end{table}

\subsection{The Try-on Parsing Inference Module}
\label{sec2.2}

Try-on semantic layout matters the try-on realism. Previous studies {\color{blue}\citep{ yang2020towards, han2018viton, wang2018toward}} have incorporated the try-on parsing inference task into garment warping and composition tasks and have implemented these tasks in one model. This design stacks the shape-level and content-level tasks together and imposes a heavy burden on model inference capacity. Following a top-down paradigm, we treat the try-on parsing inference as an individual task, which provides a more reasonable paring for downstream tasks for a more realistic result. However, the lack of ground truth for virtual try-on poses difficulties in training the try-on parsing inference, and in this section, we will describe how to tackle it in TPIM by leveraging disentanglement and consistency.

We disentangle the person parsing into three clusters: the dressed garment (${pg}$), the upper skin (${ps}$), and the remaining parts (${pr}$). When trying on a new garment, it is imperative to accurately infer the upper skin shape both when it is covered by the garment and when it is not. Therefore, we have specifically disentangle the upper skin from other categories. On this basis, we represent the remaining parts and upper skin clusters with ${{{\cal M}_{pr}}}$ and ${{\cal P}}$, input them along with the new garment mask ${{{\cal M}_g}}$ into TPIM, and then re-entangle them to infer a try-on parsing. To be specific, the person dense pose ${{\cal P}}$ is predicted by DensePose {\color{blue}\citep{guler2018densepose}}, which disentangles the shape and pose features of the upper skin from the person and disambiguates noises from the dressed garment; The person parsing ${{{\cal M}_p}}$ is predicted by Grapy-ML {\color{blue}\citep{he2020grapy}} conditioned on the person image ${{{\cal I}_p}}$, where ${{{\cal M}_p}}$ has the background, hair, face, upper skin, upper garment, leg, and lower garment categories. ${{{\cal M}_{pr}}}$ is yielded by removing the garment, upper skin, and background categories from ${{{\cal M}_p}}$.

\begin{figure}[!ht]
\centering
\includegraphics[scale=.18]{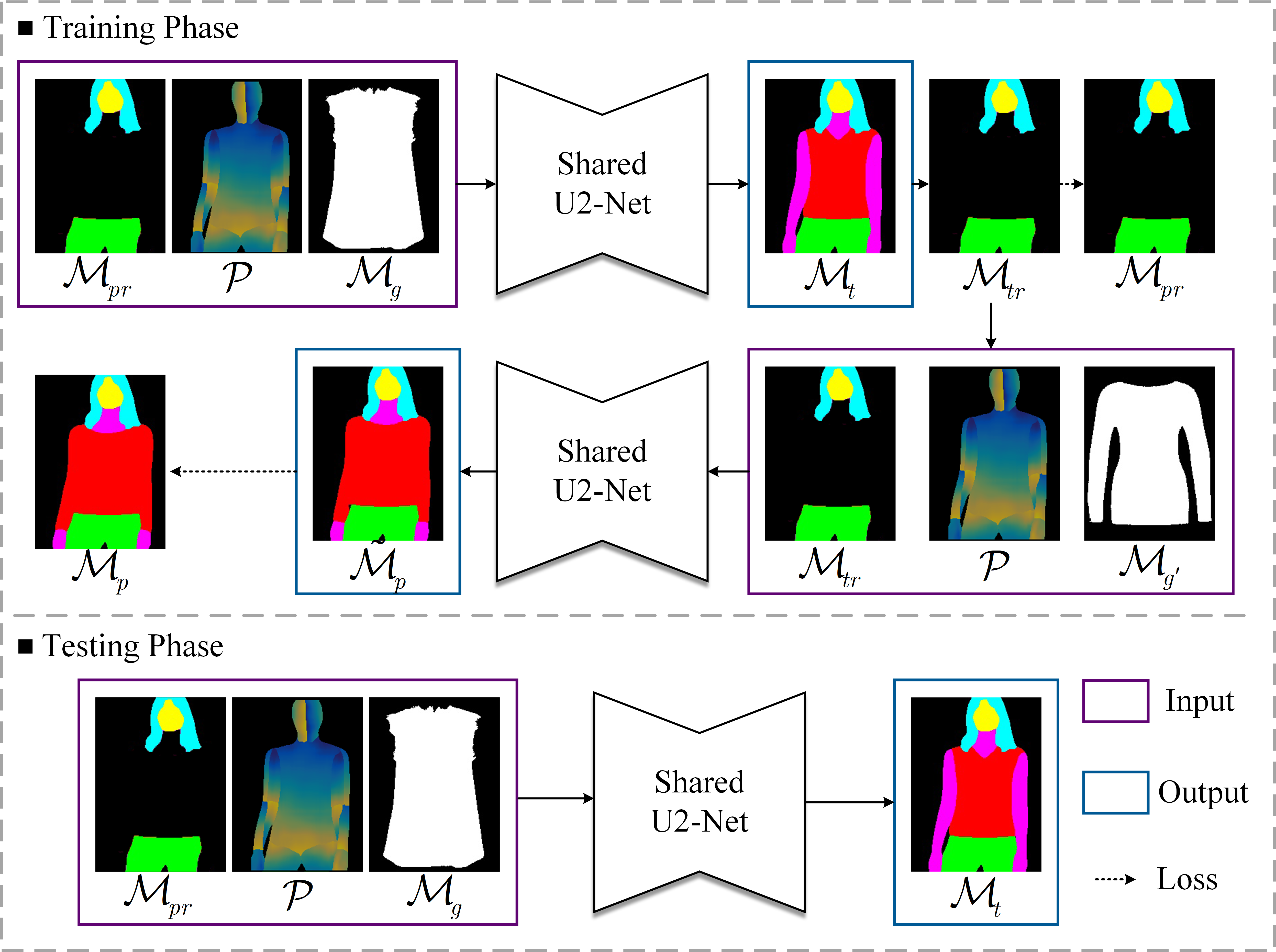}
\caption{The try-on parsing inferring module. The representations of the remaining part, upper skin, and garment are fed into U2-Net to predict a try-on parsing.}
\label{fig2}
\end{figure}

Since public datasets {\color{blue}\citep{han2018viton, wang2018toward}} collected a person image and a paired garment, previous studies {\color{blue}\citep{ yang2020towards, han2018viton, wang2018toward}} trained models by removing the dressed garment from the person and putting on this paired garment again. This training paradigm enables models to not be robust when wearing new garments during testing. In contrast, we endeavor to make the input conditions similar during the training and testing phases, and exploit the cycle consistency {\color{blue}\citep{ge2021disentangled}} to address the supervision problem caused by the ground truth lacking. Specifically, as {\color{blue} Fig. \ref{fig2}} shows, we first wear a new unpaired garment on the person, namely, TPIM infers ${{{\cal M}_t}}$ conditioned on ${{{\cal M}_{pr}}}$, ${{\cal P}}$, ${{{\cal M}_g}}$. Then, we intend to wear the paired garment ${{{\cal M}_{g'}}}$ on the preceding try-on parsing to obtain the person parsing again, namely TPIM infers a predicted person parsing ${{\tilde {\cal M}_p}}$ conditioned on ${{{\cal M}_{tr}}}$, ${{\cal P}}$, ${{{\cal M}_{g'}}}$. As for supervision, we devise the objective function of TPIM as {\color{blue} (\ref{eq1})}, where ${{\ell _1}}$ are the L1 loss; ${{\lambda _1} - {\lambda _2}}$ are weights. The former item semi-supervises ${{{\cal M}_{tr}}}$ with ${{{\cal M}_{pr}}}$, and the latter item full-supervises ${{\tilde {\cal M}_p}}$ with ${{{\cal M}_p}}$. Our training paradigm enables TPIM to learn how to entangle garment and person features and facilitates it to infer a reason parsing when trying on a new garment or wearing the paired garment.

\begin{equation}
\label{eq1}{\ell _{TPIM}} = {\lambda _1}{\ell _1}\left( {{{\cal M}_{tr}},{{\cal M}_{pr}}} \right) + {\lambda _2}{\ell _1}\left( {{{\tilde {\cal M}}_p},{{\cal M}_p}} \right)
\end{equation}

TPIM consists of two subtasks: mapping ${{{\cal M}_{pr}}}$ into ${{{\cal M}_{tr}}}$ and re-entangling upper skin features with garment features. To fulfill the high requirements for low-level feature transferring and high-level feature understanding, we adopt U2-Net {\color{blue}\citep{qin2020u2}} as the backbone of TPIM. U2-Net is capable of perceiving context information by fusing learned multiscale features and is suitable for the aforementioned task. We concatenate all inputs at the beginning and fed them into U2-Net, then U2-Net predicts results with seven channels through the Softmax layer.

\subsection{The Progressive Try-on Module}
\label{sec2.3}

\subsubsection{The Progressive Try-on Insight}
\label{sec2.3.1}

\begin{figure}[!ht]
\centering
\includegraphics[scale=.175]{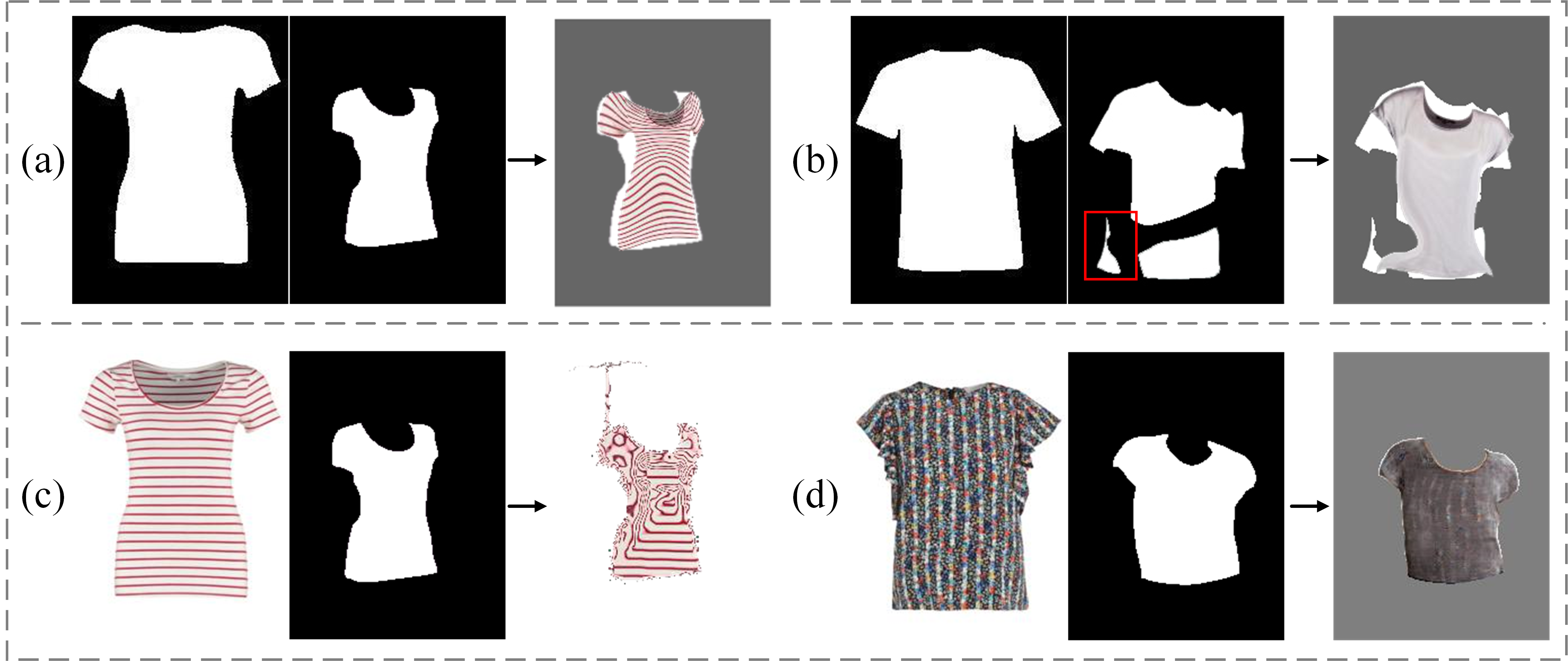}
\caption{The failure case of virtual try-on. (a) (b) The TPS deformation; (c) The optical flow warping; (d) Only Mapping.}
\label{fig3}
\end{figure}

Trying on a new garment aims to match its shape to a target shape while maintaining its style. Previous studies have leveraged TPS deformation {\color{blue}\citep{bookstein1989principal}} to downsample images into coarse grids, and then aligned grid control points to warp the images. {\color{blue} Fig. \ref{fig3}(a)} shows that its warped result does not completely match the target shape due to low-DOF warping. Despite the development of high-DOF warping tools, such as optical flow {\color{blue}\citep{dosovitskiy2015flownet}}, simple warping techniques are still inadequate in handling the try-on case, as shown in {\color{blue} Fig. \ref{fig3}(b)} where the garment is split by the arm. As a result, it is imperative to utilize feature mapping to infer the content in the red box region. To enable trying on a new garment feasible in the general case, it is imperative to employ both warping and mapping techniques. If only warping is exploited, it will result in misaligned problems for TPS and texture distortion for optical flow shown in {\color{blue} Fig. \ref{fig3}(c)}. Similarly, if only mapping is exploited, high-frequency details will be discarded during the mapping from semantic features to pixel-level content in {\color{blue} Fig. \ref{fig3}(d)}.

Therefore, we intend to warp the garment by TPS deformation to yield coarse results. And then synthesize a fined result by learning such a mapping that conforms to the coarse result at the aligned region and infers pixel-level contents at the misaligned region based on learned semantic features. However, since the fined result still suffers from details loss due to the limited spatial size of feature maps, we further composite the coarse and fined results at the pixel level to facilitate the direct propagation of local details. We name process above as the progressive try-on. In the following section, we describe how to implement the progressive try-on in PTM with the coarse warping stage, the fined mapping stage, and the composition stage.

As {\color{blue} Fig. \ref{fig4}} shows, we supervise the PTM training with the paired person and garment described in {\color{blue} Sec. \ref{sec2.2}}, namely, to match the shape of the paired garment ${{{\cal I}'_g}}$ to its dressed shape ${{{\cal M}_{pg}}}$. When the PTM has sufficient warping and mapping capabilities, in the testing phase, it is able to match the shape of a new garment ${{{\cal I}_g}}$ to the try-on shape ${{{\cal M}_{tg}}}$ inferred by TPIM.

\subsubsection{The Coarse Warping Stage}
\label{sec2.3.2}

During the coarse warping stage, we train the model to learn TPS deformation by comparing the shape difference between ${{{\cal M}_{g'}}}$ and ${{{\cal M}_{pg}}}$, enabling it to warp ${{{\cal I}_g}}$ towards ${{{\cal M}_{tg}}}$ during the testing phase. Therefore, it is imperative to devise a network to capture long-range interactions for this coarse warping, and in this section, we describe how to fulfill it by employing a ViT-based model.

\begin{figure*}[!ht]
\centering
\includegraphics[scale=.38]{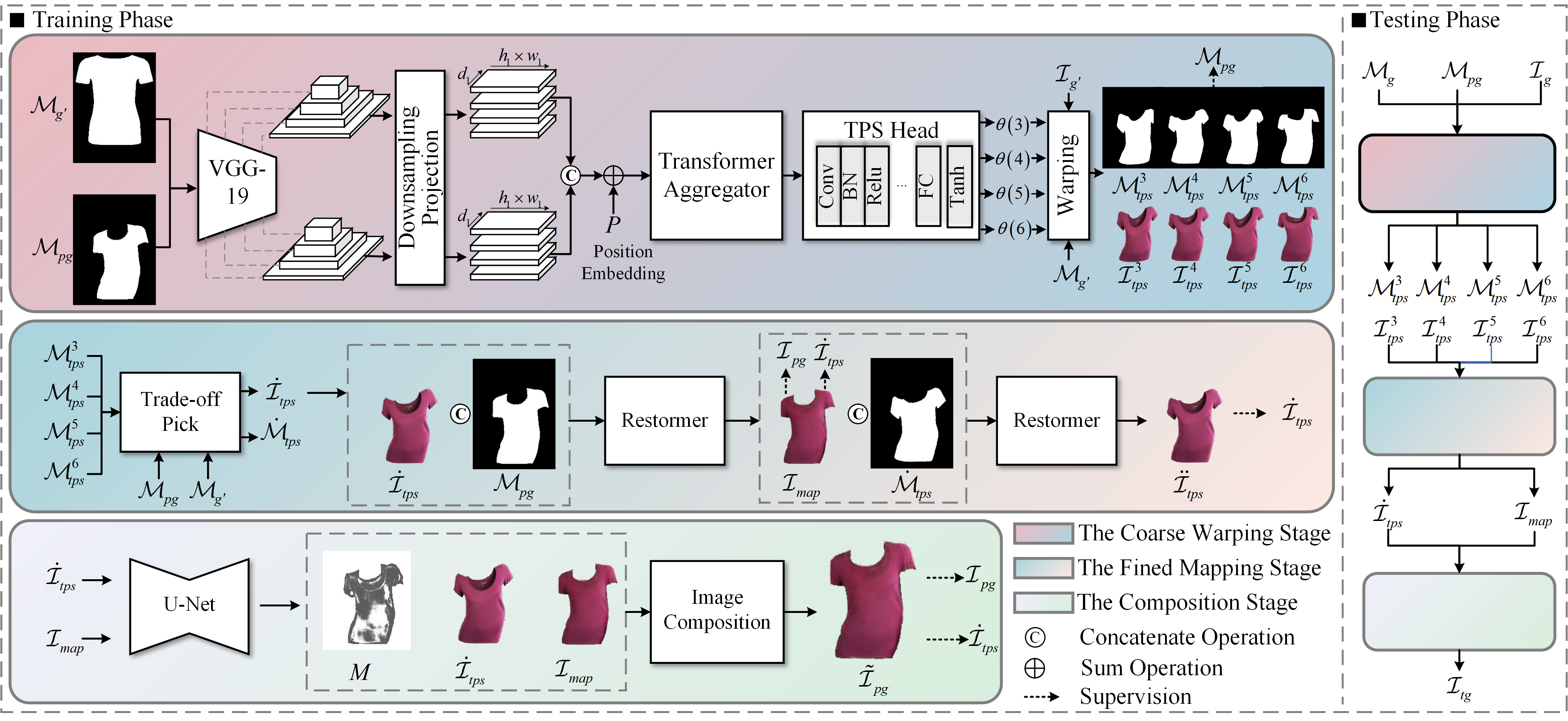}
\caption{The illustration of PTM. The coarse warping stage implements TPS deformation across four specific scales to cover a wider range of warping circumstances. The fined mapping stage selects an optimal warped result and utilizes content mapping to pad the misaligned regions. Subsequently, the composition stage concatenates the warped and mapped results at the pixel level.}
\label{fig4}
\end{figure*}

As {\color{blue} Fig. \ref{fig4}} shows, during the training phase, the coarse warping stage is conditioned on ${{{\cal M}_{g'}}}$ and ${{{\cal M}_{pg}}}$. We fed ${{{\cal M}_{g'}}}$ and ${{{\cal M}_{pg}}}$ into pre-trained VGG-19 model {\color{blue}\citep{simonyan2014very}} in turn to obtain the feature map ${\alpha _{g'}^i \in {{ \mathbb{R} } ^{{h_i} \times {w_i} \times {d_i}}}}$ of ${{{\cal M}_{g'}}}$ and the feature map ${\alpha _{pg}^i \in {{ \mathbb{R} } ^{{h_i} \times {w_i} \times {d_i}}}}$ of ${{{\cal M}_{pg}}}$. By concatenating feature maps at ${'relu1\_1'}$, ${'relu2\_1'}$, ${'relu3\_1'}$, ${'relu4\_1'}$ layers, we obtain shape feature pyramids ${{\left\{ {\alpha _{g'}^i} \right\}_{i = 1,2,3,4}}}$ and ${{\left\{ {\alpha _{pg}^i} \right\}_{i = 1,2,3,4}}}$. However, since ViT requires a constant token sequence length, we project every ${\alpha _{g'}^i}$ and ${\alpha _{pg}^i}$ into a fixed dimension ${{h_1} \times {w_1} \times {d_1}}$ to yield ${{\alpha '_{g'}},{\alpha '_{pg}} \in {{ \mathbb{R} }^{4 \times {h_1}{w_1} \times {d_1}}}}$. And then, we concatenate them and embed the position information ${P \in {{ \mathbb{R} }^{4 \times {h_1}{w_1} \times 2{d_1}}}}$ to obtain a shape patch embedding sequence ${\alpha  \in {{ \mathbb{R} }^{4 \times {h_1}{w_1} \times 2{d_1}}}}$. ${\alpha }$ is defined as {\color{blue} (\ref{eq2})} where ${\copyright}$ is the concatenate operation.

\begin{equation}
\label{eq2}
\alpha  = \left( {{{\alpha '}_{g'}}\copyright{{\alpha '}_{pg}}} \right) + P
\end{equation}

To learn token relationships and global context information for predicting the attention score ${a}$, we employ a cascade of three Transformer aggregators {\color{blue}\citep{ peebles2022gan}} with the self-attention mechanism. Rather than regressing TPS coefficients in a single scale {\color{blue}\citep{han2018viton, wang2018toward}}, we predict TPS coefficients at multiple scales exploiting the same TPS head. This approach enables the model to cover a broader range of warping circumstances without incurring significant computational costs. Specifically, we select four scales ${\theta \left( 3 \right) \in {{ \mathbb{R} }^{2 \times 3 \times 3}}}$, ${\theta \left( 4 \right) \in {{ \mathbb{R} }^{2 \times 4 \times 4}}}$, ${\theta \left( 5 \right) \in {{ \mathbb{R} }^{2 \times 5 \times 5}}}$, ${\theta \left( 6 \right) \in {{ \mathbb{R} }^{2 \times 6 \times 6}}}$, which warp the paired garment image ${{{\cal I}_{g'}}}$ into ${{\cal I}_{tps}^3}$, ${{\cal I}_{tps}^4}$, ${{\cal I}_{tps}^5}$, ${{\cal I}_{tps}^6}$, and warp the garment mask ${{{\cal M}_{g'}}}$ into ${{\cal M}_{tps}^3}$, ${{\cal M}_{tps}^4}$, ${{\cal M}_{tps}^5}$, ${{\cal M}_{tps}^6}$.

\subsubsection{The Fined Mapping Stage}
\label{sec2.3.3}

\begin{figure}[!ht]
\centering
\includegraphics[scale=.18]{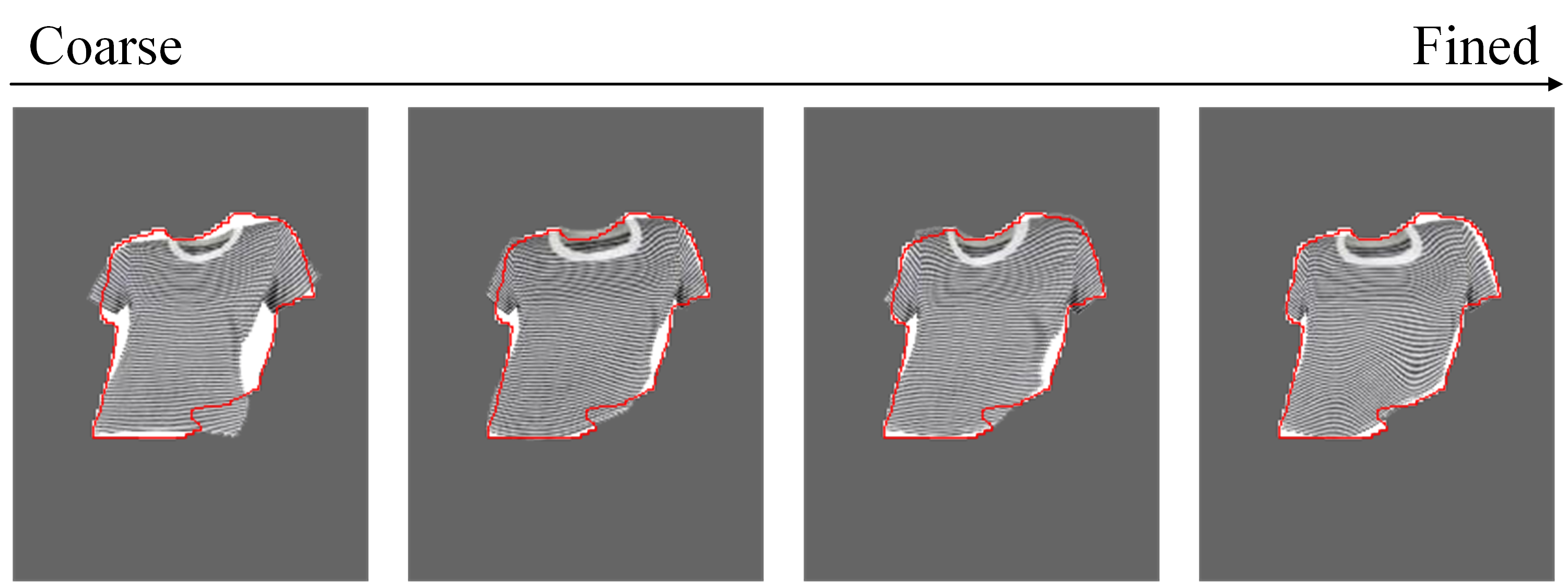}
\caption{The warping extent and texture reservation in multiscale TPS deformation.}
\label{fig5}
\end{figure}

Since TPS deformation meshes statically, the shape similarity of each scale between the target shape and the warped result varies depending on the specific virtual try-on circumstance. To establish a unique reference for the subsequent stages, we select the most suitable one from four coarse results. By this means, the coarse warping stage learns fixed four scales of TPS deformation to cover the general try-on circumstances as comprehensively as possible, and subsequently, the fined mapping stage dynamically selects an optimal one from them as a reference. We call this process "covering general and select optimal". The dynamical selection follows two rules: 1) the coarse result should be similar enough to the target shape, otherwise, it is a thorny problem to learn content mapping in their large misaligned region during the fined mapping stage; 2) the coarse result should have better content and texture reservation. Intuitively, the warping extent conflicts with the texture reservation. As shown in {\color{blue} Fig. \ref{fig5}}, the former prefers fine meshing while the latter prefers coarse meshing. Therefore, during the training phase, we compare warped results for each scale with the target shape and its source shape to select an optimal one that satisfies {\color{blue} (\ref{eq3})}. During the testing phase, we replace ${{{\cal M}_{g'}}}$ with ${{{\cal M}_g}}$.

\begin{equation}
\label{eq3}
\resizebox{1.0\hsize}{!}{${\mathop {\tilde i = \arg \min }\limits_i \left[ {\tau  \cdot avg\left( {\left\| {{\cal M}_{tps}^i - {{\cal M}_{pg}}} \right\|} \right) + \left( {1 - \tau } \right) \cdot avg\left( {\left\| {{\cal M}_{tps}^i - {{\cal M}_{g'}}} \right\|} \right)} \right]}$}
\end{equation}

\noindent where ${\tilde i}$ is the scale index of the optimal coarse result; ${\tau }$ is the trade-off weight and equals to 0.2 in our case; ${avg\left(  \cdot  \right)}$ is the average function along the spatial dimension. Thus, the optimal coarse shape ${{\dot {\cal M}_{tps}}}$ is ${{\cal M}_{tps}^{\tilde i}}$, and the optimal coarse image ${{\dot {\cal I}_{tps}}}$ is ${{\cal I}_{tps}^{\tilde i}}$.

Restormer {\color{blue}\citep{ zamir2022restormer}} is employed as the backbone for the fined mapping stage in consideration of its integrated GAN architecture and Transformer mechanism. To sufficiently acquire the mapping ability from semantic features to pixel-level content during the training phase, the cycle consistency is incorporated to facilitate the training of the fined mapping stage. To be specific, we train the Restormer to map pixel-level contents specifically in the misaligned region attributed to ${{{\cal M}_{pg}}}$ but not to ${{\dot {\cal M}_{tps}}}$. Thus, we fed ${{\dot {\cal I}_{tps}}}$ and ${{{\cal M}_{pg}}}$ into the Restormer to predict a mapped result ${{{\cal I}_{map}}}$. To prevent the content mapping is implemented at the overall shape of ${{{\cal M}_{pg}}}$, ${{\dot {\cal I}_{tps}}}$ and ${{{\cal I}_{pg}}}$ are exploited to supervise the prediction of ${{{\cal I}_{map}}}$, which facilitates the direct propagation of source content in the aligned region. As the content of the misaligned region attributed to ${{\dot {\cal M}_{tps}}}$ but not in ${{{\cal M}_{pg}}}$ is discarded during the aforementioned process, we follow the same approach, feeding ${{{\cal I}_{map}}}$ and ${{\dot {\cal M}_{tps}}}$ into the same Restormer to predict ${{\ddot {\cal I}_{tps}}}$, and supervising this prediction with ${{{\cal I}_{map}}}$ and ${{\dot {\cal I}_{tps}}}$.

During the testing phase, we feed ${{\dot {\cal I}_{tps}}}$ and ${{{\cal M}_{tg}}}$ into the well-trained Restormer once to predict ${{{\cal I}_{map}}}$.

\subsubsection{The Composition Stage}
\label{sec2.3.4}
The content mapping at the misaligned region suffers from information loss due to the spatial size limitation of semantic features. Additionally, it is inevitable to introduce the high-frequency details loss in mapping from high-level semantic features to low-level pixel contents. To address these issues, we propose to directly propagate textures and details from the optimal coarse result to the final result at the pixel level, leveraging a composition mask as a propagation carrier {\color{blue}\citep{wang2018toward}}.

Specifically, we utilized U-Net {\color{blue}\citep{ronneberger2015u}} to predict a composition mask ${M}$ conditioned on ${{\dot {\cal I}_{tps}}}$ and ${{{\cal I}_{map}}}$ through the normalization layer Sigmoid. During the training phase, guided by ${M}$, we composite ${{\dot {\cal I}_{tps}}}$ and ${{{\cal I}_{map}}}$ to yield the final result ${{\tilde {\cal I}_{pg}}}$ as {\color{blue} (\ref{eq4})} where ${ \odot }$ is the element-wise multiplication. The testing phase follows the same paradigm to obtain the final result ${{{\cal I}_{tg}}}$.

\begin{equation}
\label{eq4}
{\tilde {\cal I}_{pg}} = {{\cal I}_{map}} \odot M + {\dot {\cal I}_{tps}} \odot \left( {1 - M} \right)
\end{equation}

\subsubsection{Objective function}
\label{sec2.3.5}
We conduct joint training of three stages exploiting paired person and garment data. As these stages vary in task and objective, the error is only propagated within each stage, and the outputs of the previous stage are cloned and detached to serve as inputs for the subsequent stage. This training strategy facilitates error convergence in optimizing models.

The objective function of the coarse warping stage ${{\ell _{cw}}}$ is formulated as {\color{blue} (\ref{eq5})} where ${{\ell _1}}$ and ${{\ell _2}}$ are L1 and L2 losses; ${{\lambda _3}}$ and ${{\lambda _4}}$ are weights. Ground truth ${{{\cal M}_{pg}}}$ supervises the warped garment shape for each scale. Additionally, we define the source static grid and warped grid at the ${i}$-th scale are ${{{\cal G}^i} \in {{ \mathbb{R} } ^{2 \times i \times i}}}$ and ${{\cal G}_{tps}^i \in {{ \mathbb{R} } ^{2 \times i \times i}}}$ in TPS deformation. ${{{\cal G}^i}}$ and ${{\cal G}_{tps}^i}$ represent the position information of control point matrixes with the size of ${i \times i}$, i.e. ${x}$ and ${y}$ coordinate value. To prevent grid distortion, we introduce a regularization term to punish the grid movement between ${{{\cal G}^i}}$ and ${{\cal G}_{tps}^i}$.

\begin{equation}
\label{eq5}
{\ell _{cw}} = {\lambda _3}\sum\limits_{i = 3}^6 {{\ell _1}\left( {{\cal M}_{tps}^i,{{\cal M}_{pg}}} \right)}  + {\lambda _4}\sum\limits_{i = 3}^6 {{\ell _2}\left( {{\cal G}_{tps}^i,{{\cal G}^i}} \right)}
\end{equation}

In the fined mapping stage, predicted results are supervised at both the pixel and the perceptual levels via ${{\ell _1}}$ and ${{\ell _{vgg}}}$, where ${{\ell _{vgg}}}$ is calculated through the pre-trained VGG-19 model {\color{blue}\citep{simonyan2014very}} at the layers ${'relu1\_1'}$, ${'relu2\_1'}$, ${'relu3\_1'}$, ${'relu4\_1'}$. As described in {\color{blue} Sec. \ref{sec2.3.2}}, ${{\dot {\cal I}_{tps}}}$ and ${{{\cal I}_{pg}}}$ supervise ${{{\cal I}_{map}}}$ with a trade-off weight ${\xi }$ (set to 0.3 in our implementation), while ${{\dot {\cal I}_{tps}}}$ supervises ${{\ddot {\cal I}_{tps}}}$ to ensure consistency. Thus, its objective function ${{\ell _{fm}}}$ is formulated as {\color{blue} (\ref{eq6})} where ${{\lambda _5}}$ and ${{\lambda _6}}$ are weights.

\begin{equation}
\label{eq6}
\resizebox{1.0\hsize}{!}{
${\begin{array}{l}
{\ell _{fm}} = {\lambda _5}\left[ {\xi  \cdot {\ell _1}\left( {{{\cal I}_{map}},{{\dot {\cal I}}_{tps}}} \right) + \left( {1 - \xi } \right) \cdot {\ell _1}\left( {{{\cal I}_{map}},{{\cal I}_{pg}}} \right) + {\ell _1}\left( {{{\ddot {\cal I}}_{tps}},{{\dot {\cal I}}_{tps}}} \right)} \right]\\
 + {\lambda _6}\left[ {\xi  \cdot {\ell _{vgg}}\left( {{{\cal I}_{map}},{{\dot {\cal I}}_{tps}}} \right) + \left( {1 - \xi } \right) \cdot {\ell _{vgg}}\left( {{{\cal I}_{map}},{{\cal I}_{pg}}} \right) + {\ell _{vgg}}\left( {{{\ddot {\cal I}}_{tps}},{{\dot {\cal I}}_{tps}}} \right)} \right]
\end{array}}$
}
\end{equation}

In the composition stage, we supervise the content composition at the aligned region with ${{\dot {\cal I}_{tps}}}$ and ${{{\cal I}_{pg}}}$ to facilitate the direct propagation of pixel contents. Additionally, we supervise the one at the misaligned region by ${{{\cal I}_{pg}}}$. Thus, the objective function ${{\ell _{cp}}}$ is as {\color{blue} (\ref{eq7})}.

\begin{equation}
\label{eq7}
\resizebox{1.0\hsize}{!}{${{\ell _{cp}} = \xi  \cdot \left( {{{\cal M}_{pg}} \cap {{\dot {\cal M}}_{tps}}} \right) \cdot {\ell _1}\left( {{{\tilde {\cal I}}_{pg}},{{\dot {\cal I}}_{tps}}} \right) + \left( {1 - \xi } \right) \cdot {\ell _1}\left( {{{\tilde {\cal I}}_{pg}},{{\cal I}_{pg}}} \right)}$
}
\end{equation}

\subsection{ Re-naked Skin Inpainting Module}
\label{sec2.4}

When trying on a new short-sleeve garment on an imaged person dressed in a long-sleeve garment, skin regions such as the arm or neck may become exposed, it is imperative to infer the exposed content for a better try-on result, which we refer to as the re-naked skin inpainting. Previous studies {\color{blue}\citep{ han2018viton, wang2018toward}} have incorporated this task in the same model as the content mapping of the garment, but this poses a heavy burden on the model due to the widely varying semantic features of the garment and skin, resulting in unrealistic skin inpainting. Conversely, when trying on a long-sleeve garment on a person dressed in a long-sleeve garment, we only cover parts of skins via Boolean operation. Previous studies {\color{blue}\citep{ge2021disentangled}} have combined skin inpainting and covering into a single task, this strategy enables the model to be vague about whether and where to inpainting during the testing phase. Therefore, in this section, we describe how to explicitly train the model to conduct the skin inpainting task via StyleGAN2 in a self-supervision manner.

\begin{figure*}[!ht]
\centering
\includegraphics[scale=.28]{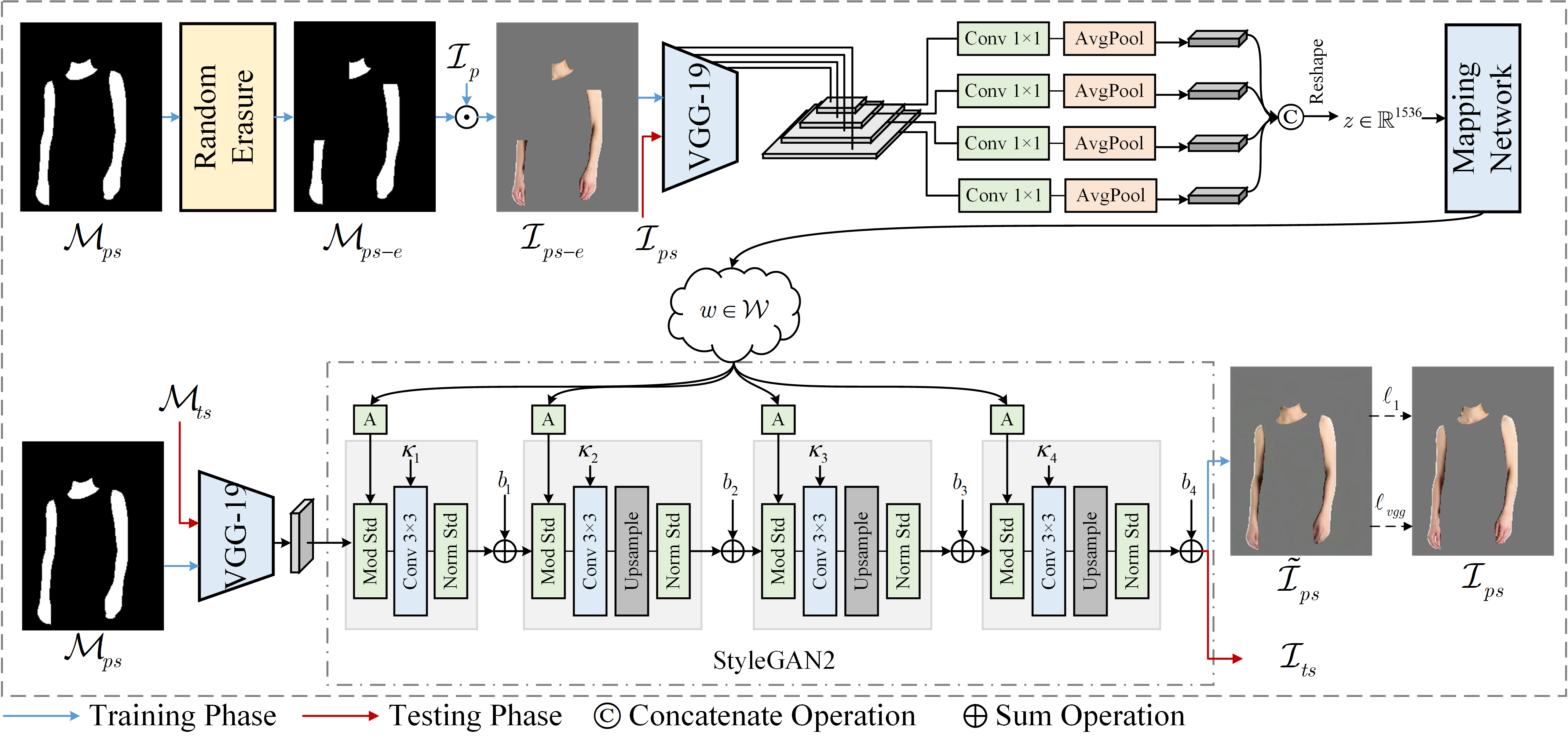}
\caption{The re-naked skin inpainting module. During the training phase, RSIM is conditioned on the erased ${{{\cal I}_{ps - e}}}$ and target shape ${{{\cal M}_{ps}}}$ to predict ${{\tilde {\cal I}_{ps}}}$, supervising with ${{{\cal I}_{ps}}}$. During the testing phase, RSIM can infer ${{{\cal I}_{ts}}}$ with the conditions of ${{{\cal I}_{ps}}}$ and ${{{\cal M}_{ts}}}$.}
\label{fig6}
\end{figure*}

During the training phase, we random erase {\color{blue}\citep{ zhong2020random}} the upper skin mask ${{{\cal M}_{ps}}}$ to form the erased mask ${{{\cal M}_{ps - e}}}$ and multiply it with ${{{\cal I}_p}}$ to yield the erased ${{{\cal I}_{ps - e}}}$, RSIM is trained to inpaint the shape ${{{\cal M}_{ps}}}$ with the content prior of ${{{\cal I}_{ps - e}}}$. Specifically, as {\color{blue} Fig. \ref{fig6}} shows, we feed ${{{\cal I}_{ps - e}}}$ into a  pre-trained VGG-19 model {\color{blue}\citep{simonyan2014very}} to extract its content features ${{\left\{ {{\eta _i}} \right\}_{i = 1,2,3,4}}}$ at the layers ${'relu1\_1'}$, ${'relu2\_1'}$, ${'relu3\_1'}$, ${'relu4\_1'}$. We reduce the channel number of ${{\eta _i}}$ by a conv${1 \times 1}$ layer and resize it to ${{\eta '_i} \in {{ \mathbb{R} }^{{c_3} \times 1 \times 1}}}$ by an AveragePooling layer. All ${{\eta '_i}}$ are concatenated together and are flattened to a 1d latent code ${z \in {{ \mathbb{R} }^{1536}}}$. We further input ${z}$ into a mapping network composed of eight FC layers to extract a spatial-agnostic intermediate vector ${w \in {{ \mathbb{R} }^{1536}}}$.

StyleGAN2 takes a constant as the initial input and transfers ${w}$ into the subsequent constant feature by a learned affine transform and AdaIN. In contrast, we intend to transfer ${w}$ into a target spatial to achieve spatial remodeling. Therefore, we input ${{{\cal M}_{ps}}}$ into the same VGG-19 model to extract the output ${\nu  \in {{ \mathbb{R} }^{{c_3} \times {h_3} \times {w_3}}}}$ at the ${'relu4\_1'}$ layer, which is exploited as the initial input of StyleGAN2 instead of a constant. We discard the noise input in the StyleGAN2 and predict ${{\tilde {\cal I}_{ps}}}$ is through four AdaIN and three upsampling. In addition, we supervise ${{\tilde {\cal I}_{ps}}}$ with ${{{\cal I}_{ps}}}$ at both the pixel and perceptual level as {\color{blue} (\ref{eq8})} where ${{\lambda _7}}$ and ${{\lambda _8}}$ are weights.

\begin{equation}
\label{eq8}
{\ell _{RSIM}} = {\lambda _7}{\ell _1}\left( {{{\tilde {\cal I}}_{ps}},{{\cal I}_{ps}}} \right) + {\lambda _8}{\ell _{vgg}}\left( {{{\tilde {\cal I}}_{ps}},{{\cal I}_{ps}}} \right)
\end{equation}

In the testing phase, an initial step is to determine where ${\left( {{{\cal M}_{ts}} \cup {{\cal M}_{ps}}} \right) > {{\cal M}_{ps}}}$. If the condition holds true, a re-naked skin inpainting is conducted to infer ${{{\cal I}_{ts}}}$ conditioned on ${{{\cal I}_{ps}}}$ and ${{{\cal M}_{ts}}}$. For a better try-on result, a further step is to composite ${{{\cal I}_{ts}}}$ with ${{{\cal I}_{ps}}}$ as ${\left( {{{\cal M}_{ts}} - {{\cal M}_{hs}}} \right) \odot {{\cal I}_{ts}} + {{\cal M}_{hs}} \odot {{\cal I}_{hs}}}$ to yield a final composited one.

\section{Experiment}
\label{sec3}

\subsection{Implementation Details}
\label{sec3.1}
The dataset in {\color{blue}\citep{han2018viton,wang2018toward}} is chosen for our experiments. We further cleaned it to contain 11565 and 1698 image pairs in the training set and the test set with a resolution of 256×192. To simulate the real-world scenario of trying on new garments, we randomly shuffled the pair of garment and person. All experiments were conducted on a single 3090 NVIDIA GPU by Adam optimizer. Furthermore, it is worth noting that the three individual modules were trained independently, where the respective training settings are enumerated in {\color{blue} Table \ref{table2}}, and the hyper-parameters setting makes loss terms roughly equal.

\begin{table}[!ht]
\caption{The training settings of all modules.}

\label{table2}
\centering
\begin{tabular}{cp{30pt}p{30pt}p{30pt}p{60pt}}
\toprule    &  Lr & Iterations & Batch & Hyper-parameters   \\ \midrule
                     TPIM    &  ${1 \times {e^{ - 4}}}$          & 72${k}$    & 16 &\makecell[l]{${{\lambda _1} = 2.0}$,${{\lambda _2} = 2.0}$}\\
                     PTM     &  ${2 \times {e^{ - 4}}}$          & 145${k}$   & 4  & \makecell[l]{${{\lambda _3} = 3.0}$,${{\lambda _4} = 0.3}$,\\${{\lambda _5} = 6.0}$,${{\lambda _6} = 0.2}$}\\
                     RSIM     &  ${1 \times {e^{ - 5}}}$         & 36${k}$    & 32 & ${{\lambda _7} = 6.0}$,${{\lambda _8} = 0.2}$\\ \bottomrule
\end{tabular}
\end{table}

\subsection{Ablation Study}
\label{sec3.2}

\subsubsection{The Try-on Parsing Inference Module}
\label{sec3.2.1}

\begin{figure}[!ht]
\centering
\includegraphics[scale=.15]{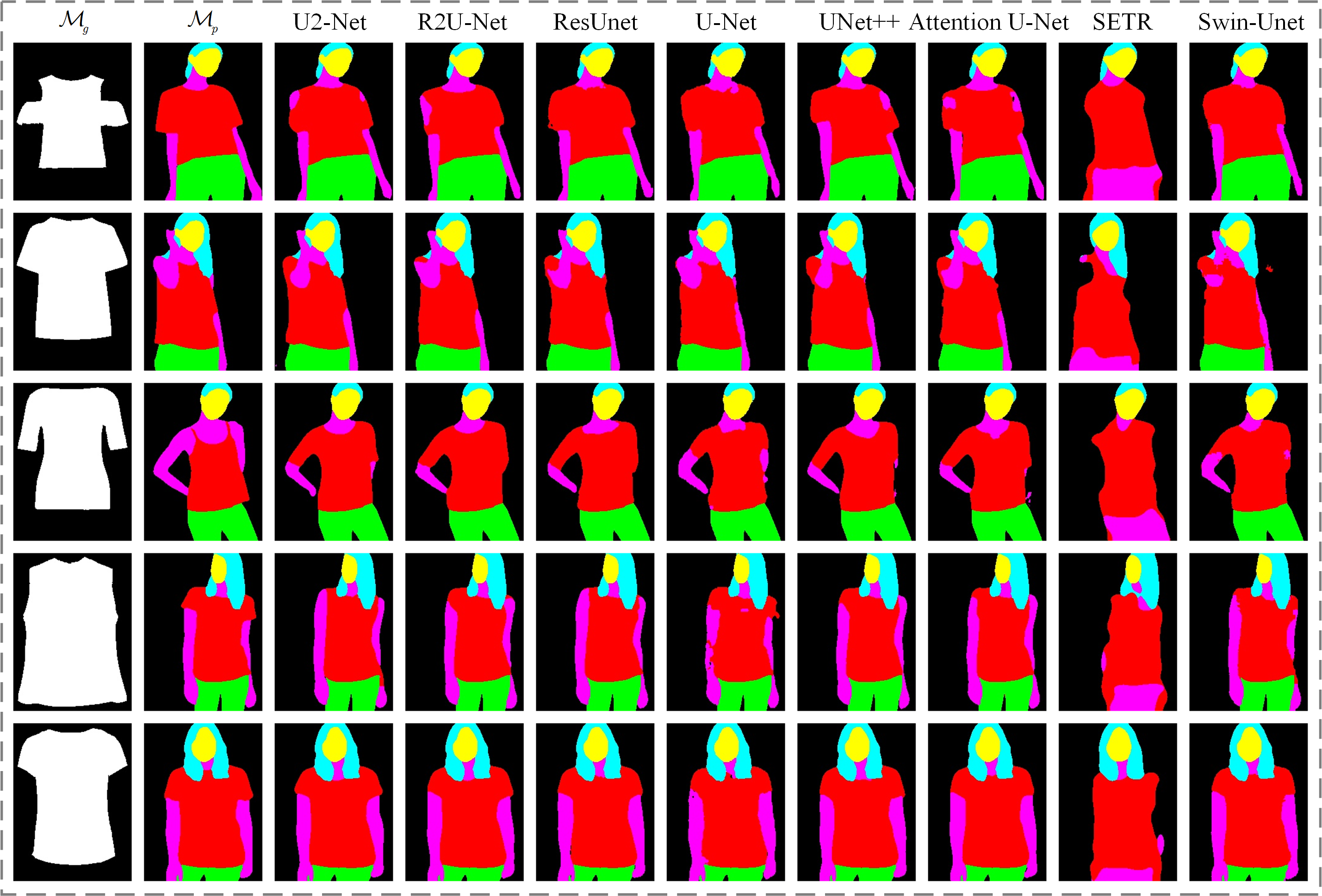}
\caption{The qualitative comparison between different backbones applied in TPIM.}
\label{fig7}
\end{figure}

We employed a diverse set of backbones to infer the try-on parsing within the TPIM framework.  They are U2-Net {\color{blue}\citep{qin2020u2}}, R2U-Net {\color{blue}\citep{alom2019recurrent}}, ResUnet {\color{blue}\citep{zhang2018road}}, U-Net {\color{blue}\citep{ronneberger2015u}}, UNet++ {\color{blue}\citep{zhou2019unet++}}, Attention U-Net {\color{blue}\citep{oktay2018attention}}, SETR {\color{blue}\citep{zheng2021rethinking}}, Swin-Unet {\color{blue}\citep{cao2021swin}}. Their qualitative comparison is shown in {\color{blue} Fig. \ref{fig7}}. When the person has a complicated pose as shown in 1-2 rows, U2-Net is capable of capturing the global context information of the person and garment resulting in reasonable try-on parsing inference. As shown in 3-5 rows, U2-Net demonstrates superior performance in generating naturally-looking local regions such as cuffs. Therefore, it demonstrates that U2-Net is more suitable to infer the reasonable try-on parsing among these backbones.

\begin{figure}[!ht]
\centering
\includegraphics[scale=.18]{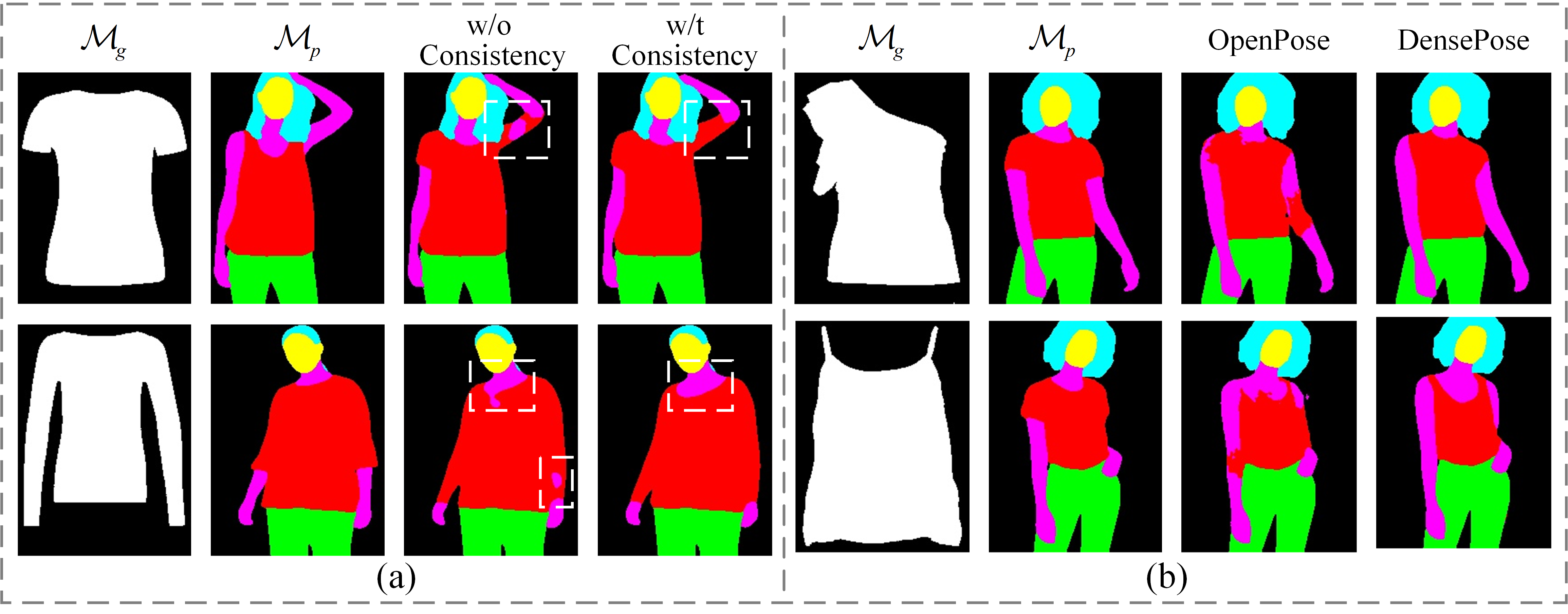}
\caption{The ablation study of TPIM. (a) The ablation of consistency; (b) The qualitative comparison between pose forms.}
\label{fig8}
\end{figure}

As {\color{blue} Fig. \ref{fig8}(a)} shows, we present the ablation study of consistency. As consistency enables TPIM aware of more try-on cases and prevents the inference process from succumbing to rigid patterns, TPIM is more robust when trying on a new garment with a style and category substantially distinct from its dressed one. As {\color{blue} Fig. \ref{fig8}(b)} shows, we have also verified the effect of pose forms on TPIM, exploiting the OpenPose method {\color{blue}\citep{cao2017realtime}} to extract the joint position of the person as a comparison. As the OpenPose method falls short in representing a person shape, it weakens the inferring capacity of TPIM at the cross-category region, causing blur at the boundary between arm and cuff.

\subsubsection{The Coarse Warping Stage}
\label{sec3.2.2}

\textbf{Baselines} This section endeavors to verify the function of grid regularization and compares ViT to CNN on learning garment warping. To be specific, we leverage the GMM module of CP-VTON {\color{blue}\citep{wang2018toward}} as the baseline of CNN, and the coarse warping stage is employed as our baseline. The experiment configurations are set to the same for a fair comparison.

\begin{figure}[!ht]
\centering
\includegraphics[scale=.18]{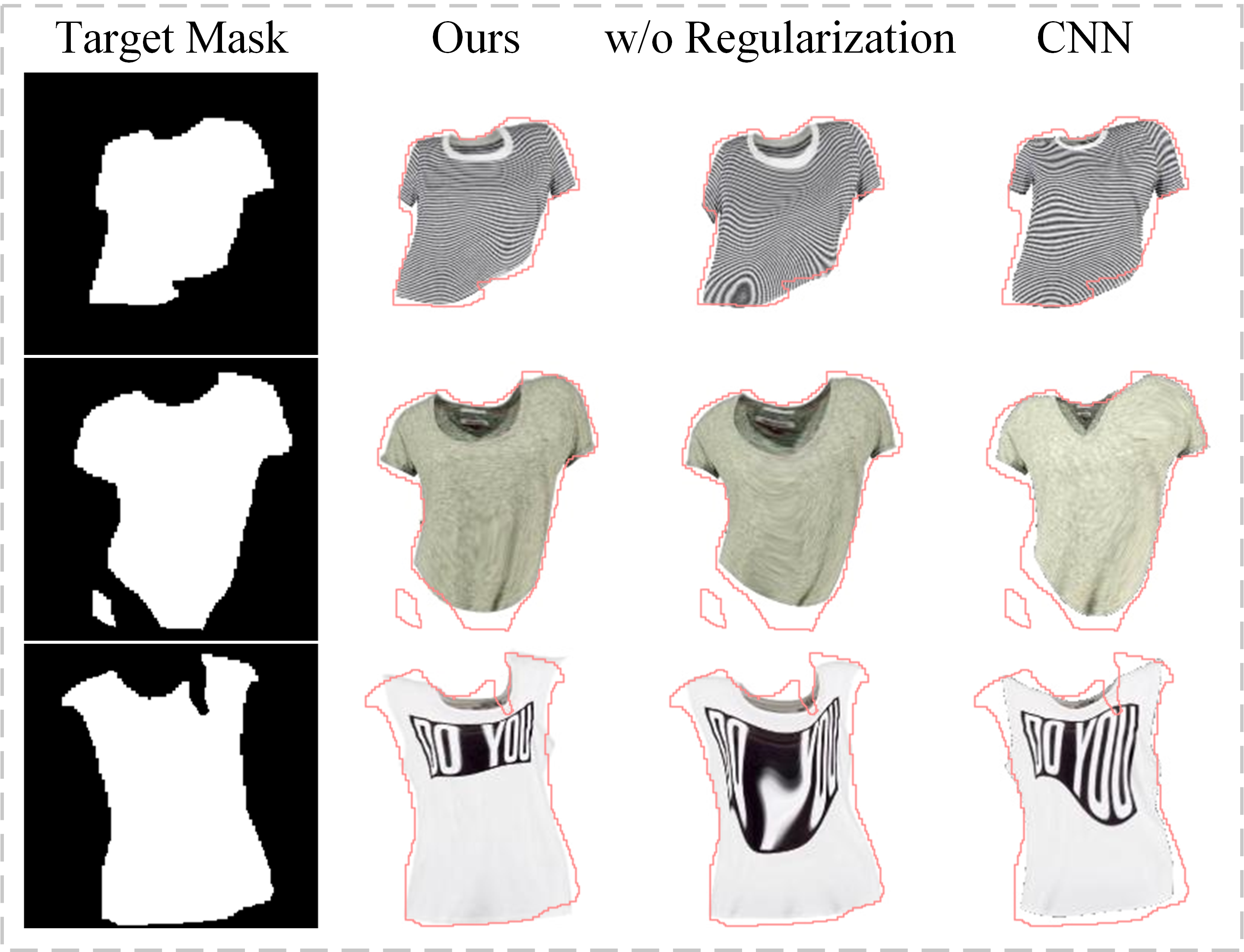}
\caption{The qualitative comparison in the coarse warping stage.}
\label{fig9}
\end{figure}

\textbf{Metrics} We evaluate baselines from two aspects: warping extent and texture reservation. The warping extent is quantified through calculating MSE between the warped garment mask and the target mask predicted by TPIM. As the warping ground truth is unavailable, we calculate the FID between warped garments and original garments to represent texture reservation. Some qualitative examples are shown in {\color{blue} Fig. \ref{fig9}}, and the quantitative results are shown in {\color{blue} Table \ref{table3}}.

\begin{table}[!ht]
\caption{The quantitative results of the coarse warping stage.}

\label{table3}
\centering
\begin{tabular}{cccc}
\toprule                                            &Ours     & w/o Regularization &CNN \\ \midrule
                     MSE${\left( {{{\cal M}_{tps}},{{\cal M}_{tg}}} \right)}$   &  2016.2            & 1873.5             & 2141.6 \\
                     FID${\left( {{{\cal I}_{tps}},{{\cal I}_g}} \right)}$               &  68.9              & 84.2               & 91.4\\ \bottomrule
\end{tabular}
\end{table}

Experiments demonstrate that grid regularization is capable of preventing texture distortion at the expense of some loss in the warping extent. Since the inner region has not a significant warping trend compared to the boundary region, the grid regularization further penalizes its random movement and suppresses distortion to promote warping coordination. Moreover, FID decreases by 18.17\% in this case. Benefiting from the long-range relationship modeling of ViT, our method performs better in learning warping garments, where our MSE${\left( {{{\cal M}_{tps}},{{\cal M}_{tg}}} \right)}$ has reduced by 12.52\% compared to CNN warping. It demonstrates our superiority in perceiving coarse-grained shape differences.

\subsubsection{The Fined Mapping Stage}
\label{sec3.2.3}

In this section, we verify the impact of trade-off weight ${\tau }$ on selecting an optimal coarse result, the influence of another trade-off weight ${\xi }$ in the training supervision, and the ablation study of consistency and the backbone.

\begin{figure}[!ht]
\centering
\includegraphics[scale=.45]{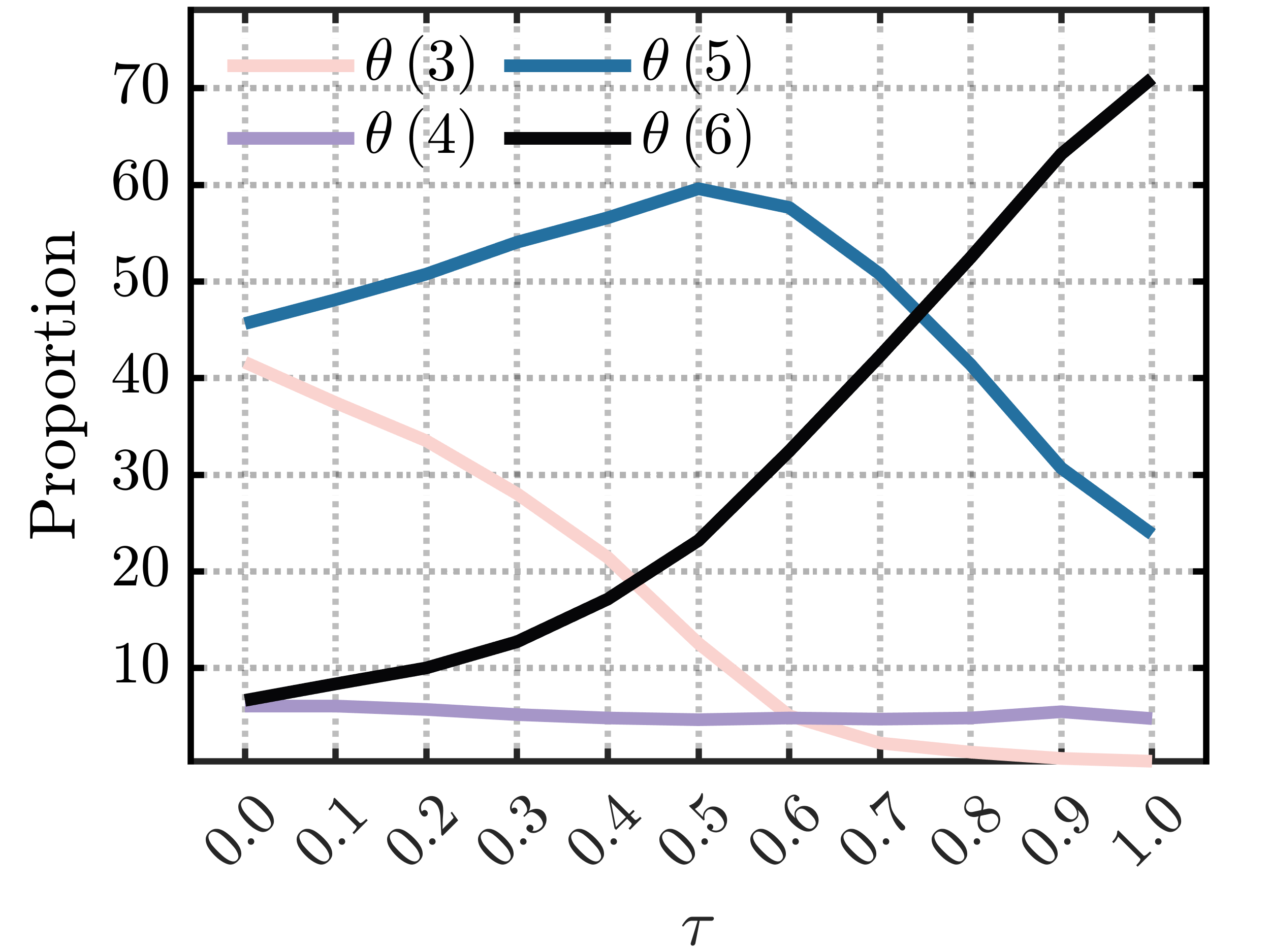}
\caption{The selecting trend with trade-off weight ${\tau }$.}
\label{fig10}
\end{figure}

As described in {\color{blue} Sec. \ref{sec2.3.3}}, we have introduced the trade-off weight ${\tau }$ to balance the warping extent and texture reservation in selecting. We reveal the selecting trend of each scale with various ${\tau }$ in {\color{blue} Fig. \ref{fig10}}.

As ${\tau }$ tends to 0, the selection prefers a coarse result with a better texture reservation, the selection proportion of scale 3 increases, while scale 6 one decreases. In contrast, as ${\tau }$ approaches 1, the selection proportion of scale 6 is increasing as a fine mesh is prone to match the shape exactly. To fulfill our intention of covering the general try-on circumstances, we expect the selection proportion of each scale to be equivalent. However, the selection proportion of scale 4 remains approximately 8\% across different ${\tau }$, we calculate the MSE of selection proportions for the other scales and find a value for ${\tau }$ with minimum MSE. Therefore, we recommend ${\tau }$ to be 0.2.

\begin{figure}[!ht]
\centering
\includegraphics[scale=.20]{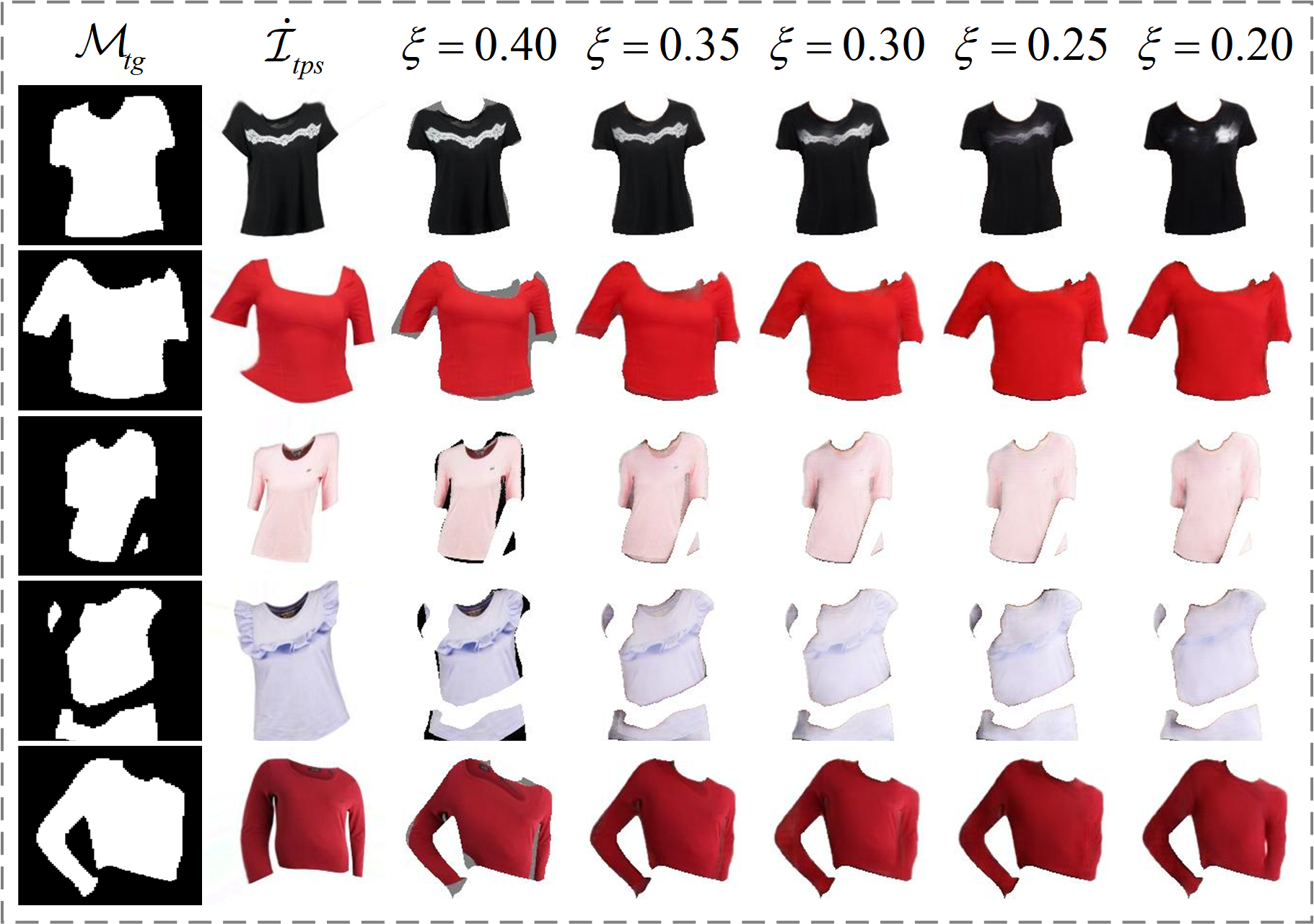}
\caption{The qualitative cases of being with different trade-off weights ${\xi }$.}
\label{fig11}
\end{figure}

As described in {\color{blue} Sec. \ref{sec2.3.3}}, we supervise ${{{\cal I}_{map}}}$ with ${{{\cal I}_{pg}}}$ and ${{\dot {\cal I}_{tps}}}$ to learn a desired mapping. We verify the influence of trade-off weight ${\xi }$ on the training from two aspects: the content transfer from ${{\dot {\cal I}_{tps}}}$ and the mapping quality, the former is represented by FID${\left( {{{\cal I}_{map}},{{\dot {\cal I}}_{tps}}} \right)}$ and SSIM${\left( {{{\cal I}_{map}},{{\dot {\cal I}}_{tps}}} \right)}$, while the latter is represented by IS {\color{blue}\citep{ salimans2016improved}}. The qualitative cases are shown in {\color{blue} Fig. \ref{fig11}}, and the quantitative trend is in {\color{blue} Table \ref{table4}}.

\begin{table}[!ht]
\caption{The mapping results with different supervision trade-off weights ${\xi }$.}
\label{table4}
\centering
\begin{tabular}{cccccc}
\toprule   ${\xi }$                                                         &0.20  & 0.25  &0.30 &0.35 &0.40 \\ \midrule
          FID${\left( {{{\cal I}_{map}},{{\dot {\cal I}}_{tps}}} \right)}$  &61.73 & 55.76 &48.71&38.85&37.86\\
          SSIM${\left( {{{\cal I}_{map}},{{\dot {\cal I}}_{tps}}} \right)}$ &0.85  & 0.86  &0.87 &0.88 &0.90 \\
          IS                                                                &3.54  & 3.48  &3.60 &3.48 &3.31\\ \bottomrule
\end{tabular}
\end{table}

As ${\xi }$ increases, ${{\dot {\cal I}_{tps}}}$ has more contribution to supervising result during the training phase, it is facilitate to transfer the content of ${{\dot {\cal I}_{tps}}}$ into ${{{\cal I}_{map}}}$ at the aligned region but suppress the content mapping at the misaligned region, resulting FID${\left( {{{\cal I}_{map}},{{\dot {\cal I}}_{tps}}} \right)}$ and SSIM${\left( {{{\cal I}_{map}},{{\dot {\cal I}}_{tps}}} \right)}$ increase, In contrast, as ${\xi }$ decreases, Restormer is prone to map contents in overall shape. Therefore, we set ${\xi }$ to 0.3 to obtain a trade-off where IS achieves a high score of 3.60.

\begin{figure}[!ht]
\centering
\includegraphics[scale=.26]{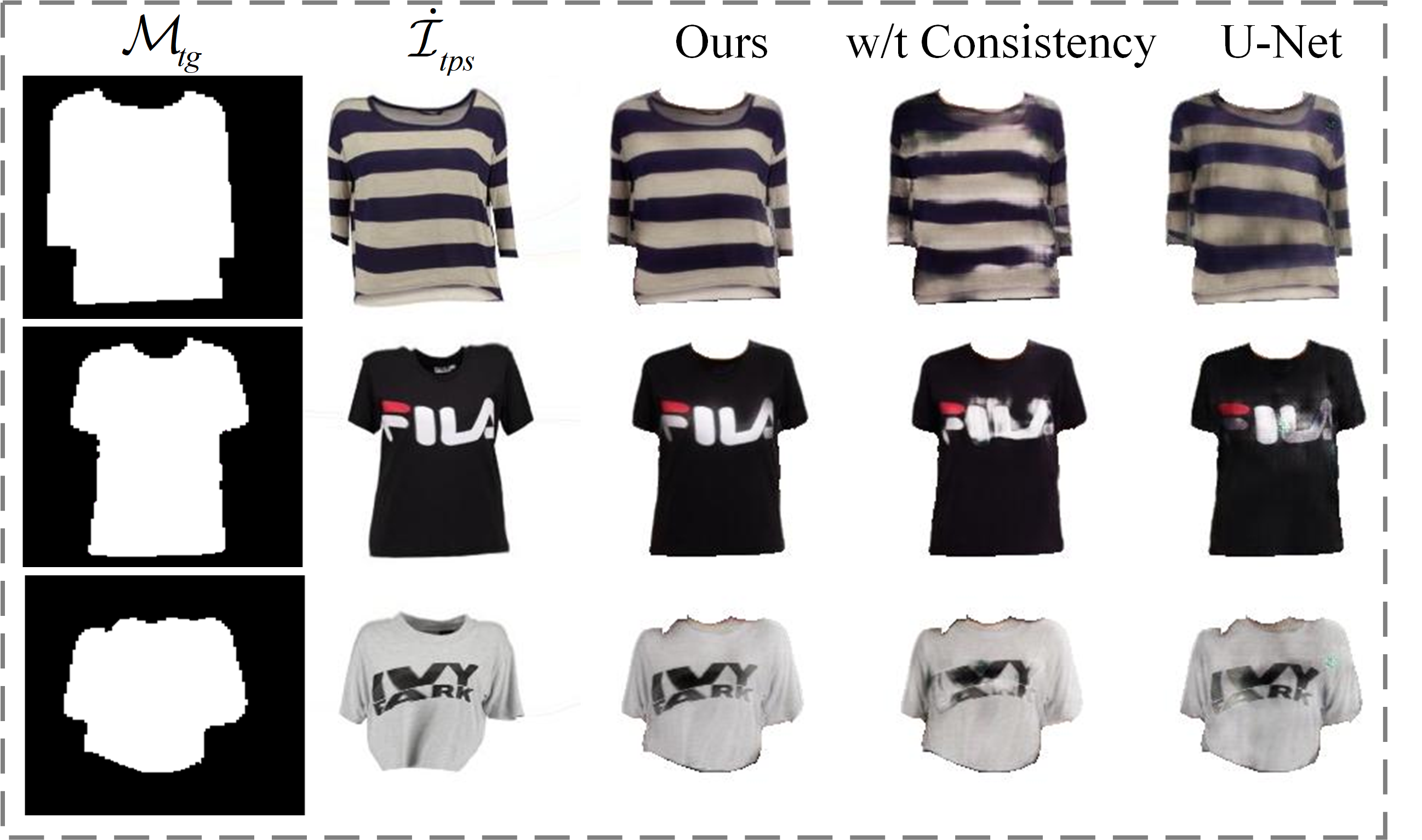}
\caption{The qualitative cases in the ablation of the fined mapping stage.}
\label{fig12}
\end{figure}

Furthermore, we verify the function of consistency and the Restormer backbone. Specifically, we employ U-Net as the backbone for a comparison, which has been leveraged in ACGPN {\color{blue}\citep{ yang2020towards}}. The qualitative and quantitative results are in {\color{blue} Fig. \ref{fig12}} and {\color{blue} Table \ref{table5}}, respectively.

\begin{table}[!ht]
\caption{The quantitative results in the ablation of the fined mapping stage.}

\label{table5}
\centering

\begin{tabular}{cccc}
\toprule
                                                                 & \makecell[c]{Ours} & \makecell[c]{w/t Consistency}   & \makecell[c]{U-Net}  \\ \hline
          FID${\left( {{{\cal I}_{map}},{{\dot {\cal I}}_{tps}}} \right)}$  &48.71 & 53.73 &75.20\\
          SSIM${\left( {{{\cal I}_{map}},{{\dot {\cal I}}_{tps}}} \right)}$ &0.87  & 0.82  &0.86  \\
          IS                                                                &3.60  & 3.60  &3.44 \\\bottomrule
\end{tabular}

\end{table}

Since consistency facilitates indirectly supervising the content propagation at the aligned region, its ablation will blur local details like the logo for ${{{\cal I}_{map}}}$ and worsen FID  and SSIM  by 10.31\% and 5.75\%. As the locality inductive bias, CNN relies on neighboring contents for mapping, rendering U-Net unsuitable for the fine mapping stage. In contrast, Restormer models the global interaction to learn content mapping, resulting in a more natural mapped result at the perception level.

\subsubsection{The Pipeline of PTM}
\label{sec3.2.4}

In this section, we verify the necessity of three stages in PTM. As described in {\color{blue} Sec. \ref{sec2.3.1}}, without the fined mapping, coarse warped results do not match the target shape exactly. Accordingly, we emphasize the ablation study of the coarse warping stage and composition stage. The image quality and texture reservation with ${{{\cal I}_g}}$ are evaluated by IS and FID${\left( {{{\cal I}_{map}},{{\cal I}_g}} \right)}$.

\begin{figure}[!ht]
\centering
\includegraphics[scale=.24]{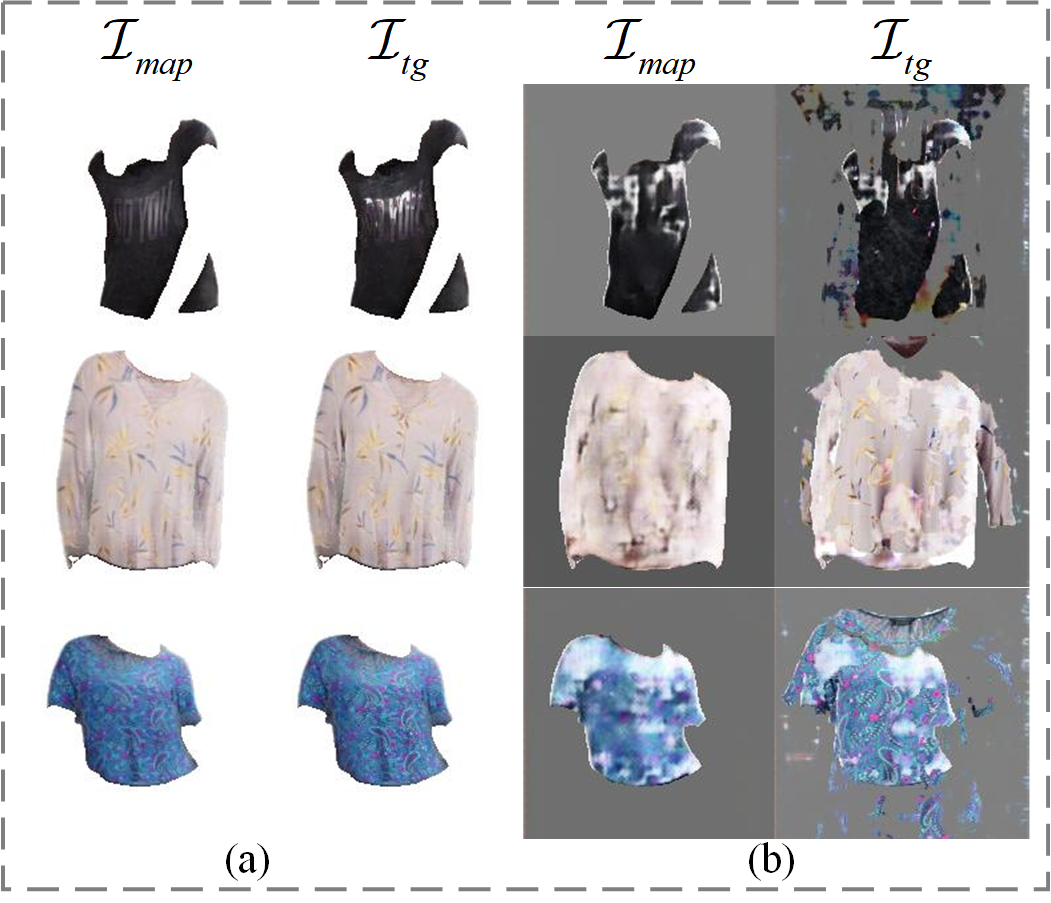}
\caption{The qualitative comparison in the ablation of pipeline. (a) Overall pipeline; (b) w/o the coarse warping stage.}
\label{fig13}
\end{figure}

\begin{table}[!ht]
\caption{The quantitative results in the ablation of pipeline.}
\label{table6}
\centering
\begin{tabular}{ccc}
\toprule
 &FID${\left( {{{\cal I}_{map}},{{\cal I}_g}} \right)}$ & IS  \\ \hline
 Overall Pipeline & 124.16 &3.59 \\
 w/o the Composition Stage & 122.17 & 3.60 \\
 w/o the Coarse Warping Stage & 295.87 & 4.47 \\\bottomrule

\end{tabular}
\end{table}

As {\color{blue} Fig. \ref{fig13}} and {\color{blue} Table \ref{table6}} show, the absence of the coarse warping stage results in a situation where the fined mapping stage is conditioned on ${{{\cal I}_g}}$ and ${{{\cal M}_{pg}}}$ to predict ${{{\cal I}_{map}}}$. The considerable shape difference between ${{{\cal M}_g}}$ and ${{{\cal M}_{pg}}}$ impedes the direct content transfer at the corresponding position, thereby enabling the model to overload at the mapping from semantic features to content pixels for the overall shape. Moreover, this absence leads to the failure of texture transfer at the pixel level in the subsequent composition stage.

For the most try-on circumstances, ${{{\cal I}_{map}}}$ exhibits satisfactory content propagation of aligned region from ${{\dot {\cal I}_{tps}}}$. In cases where this inheritance mechanism fails, the composition stage can facilitate the local details like the logo or the high-frequency information like the shadow to transfer from ${{\dot {\cal I}_{tps}}}$ to ${{{\cal I}_{map}}}$ at the pixel level. It is worth mentioning that the composition also results in some artifacts and worsens FID and IS slightly.

\subsubsection{The Re-naked Skin Inpainting Module}
\label{sec3.2.5}

We verify the impact of erasure extent ${{{\cal M}_{ps - e}}}$ during the training phase on the performance of RSIM. Concretely, we have conducted a comprehensive experiment wherein we have established nine distinct groups of erasure extents, which are numerically labeled from 1 to 9 in ascending order of erasure extent. Subsequently, we have trained RSIM using these nine erasure extents and evaluated each RSIM on ten test sets by comparing the composited result ${{\tilde {\cal I}_{ps}}}$ with ${{{\cal I}_{ps}}}$. The test set 10 has the highest erasure extent the same as the test set 9 and directly leverages the non-composited result for comparison. Its purpose is to evaluate the performance of RSIM under extreme conditions. We leverage MSE, PSNR, and SSIM as metrics.

\begin{figure}[!ht]
\centering
\includegraphics[scale=.24]{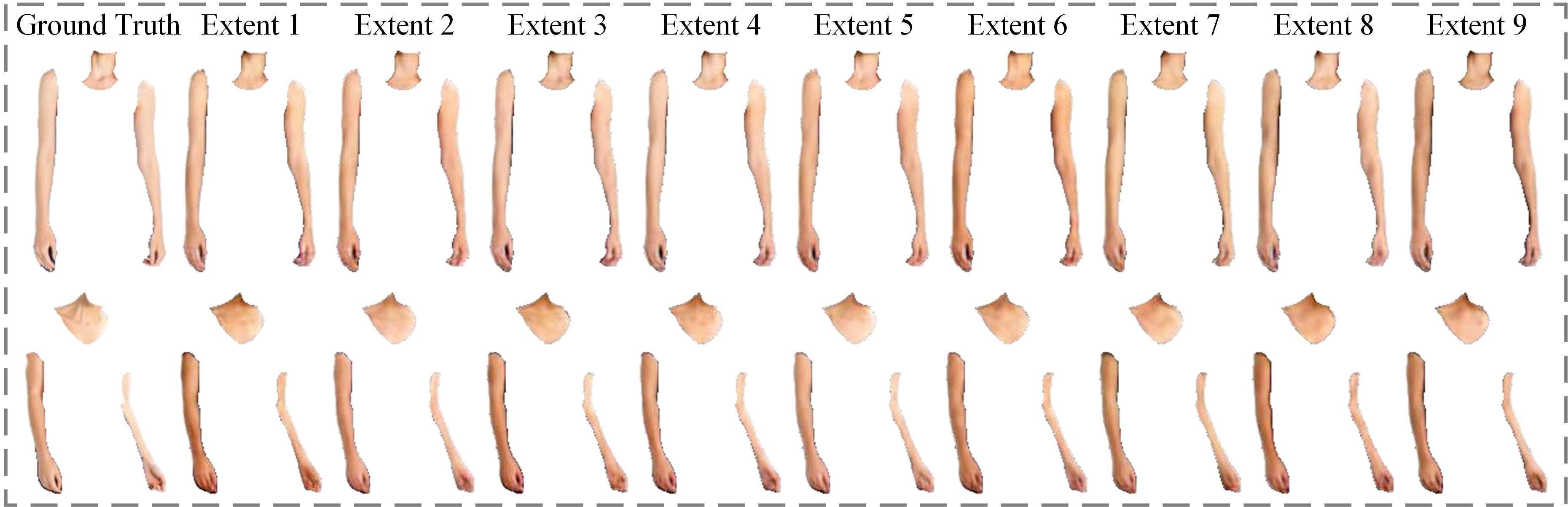}
\caption{The qualitative comparison of ${{\tilde {\cal I}_{ps}}}$ by training with various erasure extents. All examples are tested on the test set 10.}
\label{fig14}
\end{figure}

\begin{figure}[!ht]
\centering
\includegraphics[scale=.19]{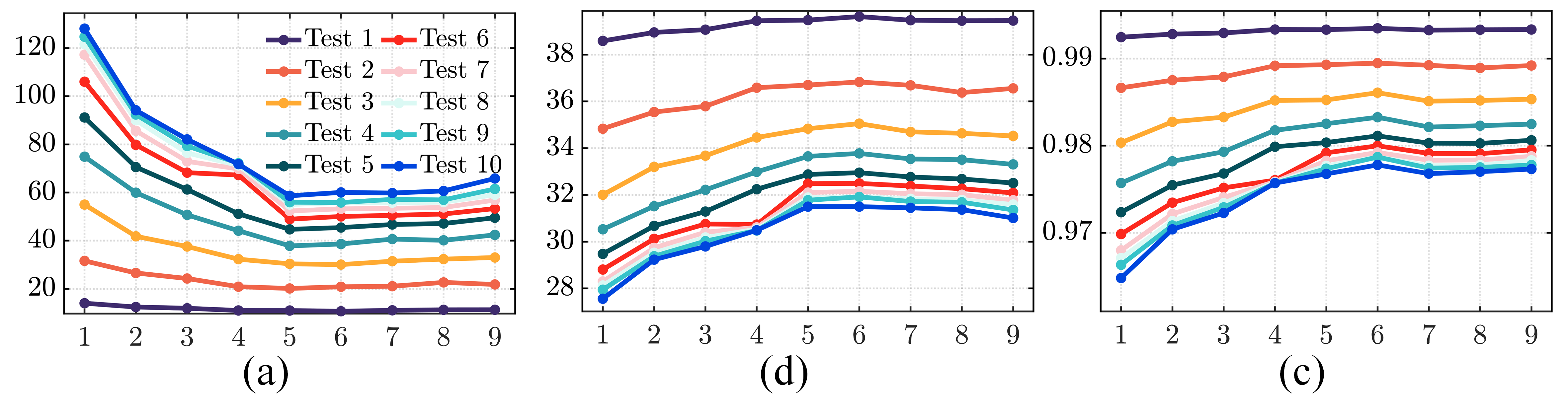}
\caption{RSIM is trained on various erasure extents and then is tested on various test sets. ${x}$-axis labels the erasure extent in the training phase, and ${y}$-axis labels (a) MSE results; (b) PSNR results; (c) SSIM results. }
\label{fig15}
\end{figure}

As {\color{blue} Fig. \ref{fig15}} shows, the performance of RSIM on all test sets improves as the erasure extent in the training phase increases to 5. However, as the erasure extent surpasses 5, the performance of RSIM degrades. This observation highlights that erasure extent 5 is the optimal choice for training RSIM, which enables it to capture the general statistical distributions of upper skin, such as arm structure, and extract specific features like skin color from ${{{\cal I}_{ps - e}}}$. Specifically, erasure extent 5 involves randomly erasing a rectangle at a position with a probability of 0.5, where the size of rectangle is ${\left[ {0.02,0.30} \right]}$ of the image area. As {\color{blue} Fig. \ref{fig14}} shows, even with the limited training condition, various trained RSIMs still predict the high-quality ${{\tilde {\cal I}_{ps}}}$ with the same style of the ground truth but are different in the details like skin color and shadow.

\begin{figure}[!ht]
\centering
\includegraphics[scale=.27]{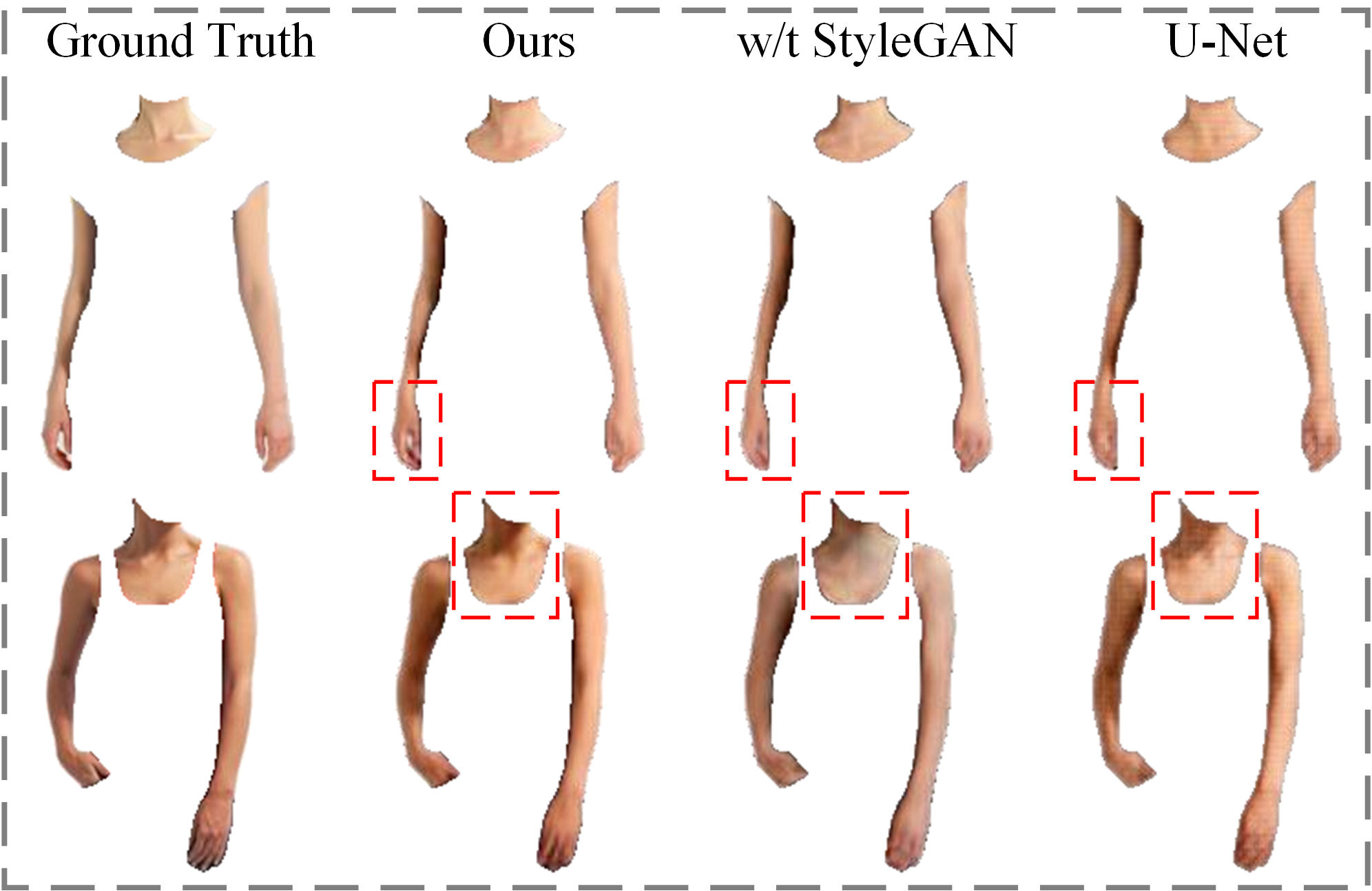}
\caption{The qualitative cases of RSIM with various backbones.}
\label{fig16}
\end{figure}

We also verify the impact of different backbones on this re-naked skin inpainting task. Concretely, we replace StyleGAN2 with its previous version StyleGAN {\color{blue}\citep{ karras2019style}} in RSIM, and leverage U-Net as RSIM only conditioned on ${{{\cal I}_{ps - e}}}$ and ${{{\cal M}_{ps}}}$. The qualitative cases are shown in {\color{blue} Fig. \ref{fig16}}, and the quantitative results are in {\color{blue} Table \ref{table7}}.

\begin{table}[!ht]
\caption{The quantitative results of RSIM.}
\label{table7}
\centering
\begin{tabular}{cccc}
\toprule
& \makecell[c]{Ours} & \makecell[c]{w/t StyleGAN }   & \makecell[c]{U-Net}  \\ \hline
         MSE  & 58.64  &112.14 &92.40\\
         PSNR & 31.50  &28.35  &29.32\\
         SSIM &0.98  &0.97   &0.97  \\
         \bottomrule
\end{tabular}
\end{table}

Since StyleGAN2 grows feature maps from coarse to fine and disentangles representation, it expert at synthesizing images with obvious context information like face and upper skin. Conversely, U-Net suffers from the low-level details representation drawback, and its predicted results exhibit undesirable characteristics, such as stripes and blurs, where MSE and PSNR worsen by 57.6\% and 7.0\%. Since the defective adaption of AdaIN in StyleGAN {\color{blue}\citep{ karras2020analyzing}}, its predicted results have artifacts and the skin color shift.

\subsection{Comparison}
\label{sec3.3}

\textbf{Baselines} We compare PG-VTON with VITON {\color{blue}\citep{han2018viton}}, CP-VTON {\color{blue}\citep{wang2018toward}}, CP-VTON+ {\color{blue}\citep{minar2020cp}}, ACGPN {\color{blue}\citep{yang2020towards}}, DCTON {\color{blue}\citep{ge2021disentangled}}, LM-VTON {\color{blue}\citep{liu2021toward}}, Flow-Style {\color{blue}\citep{he2022style}}, DAFlow {\color{blue}\citep{bai2022single}}, and PL-VTON {\color{blue}\citep{han2022progressive}}. All baselines adopt their official implementations and pre-trained models. It is worth mentioning that all baselines were trained and tested on a common dataset.

\textbf{Metrics} Since there is no ground truth in virtual try-on, it is challenging to evaluate the effect of virtual try-on. As in previous studies {\color{blue}\citep{ ge2021disentangled, liu2021toward, he2022style, bai2022single}}, we compare results with half reference and no reference metrics. Specifically, the former includes SSIM and FID between ${{{\cal I}_t}}$ and ${{{\cal I}_p}}$, while the latter includes IS {\color{blue}\citep{salimans2016improved}} and hyperIQA {\color{blue}\citep{su2020blindly}}. However, it is imperative to be concerned about the garment style reservation, upper skin presentation, and the rationality of try-on parsing, which is a hard issue to be quantized. Therefore, we endeavor to show more qualitative examples in diverse situations and assess results by our perception, and conducted the human subjective assessment experiments.

\begin{figure}[!ht]
\centering
\includegraphics[scale=.23]{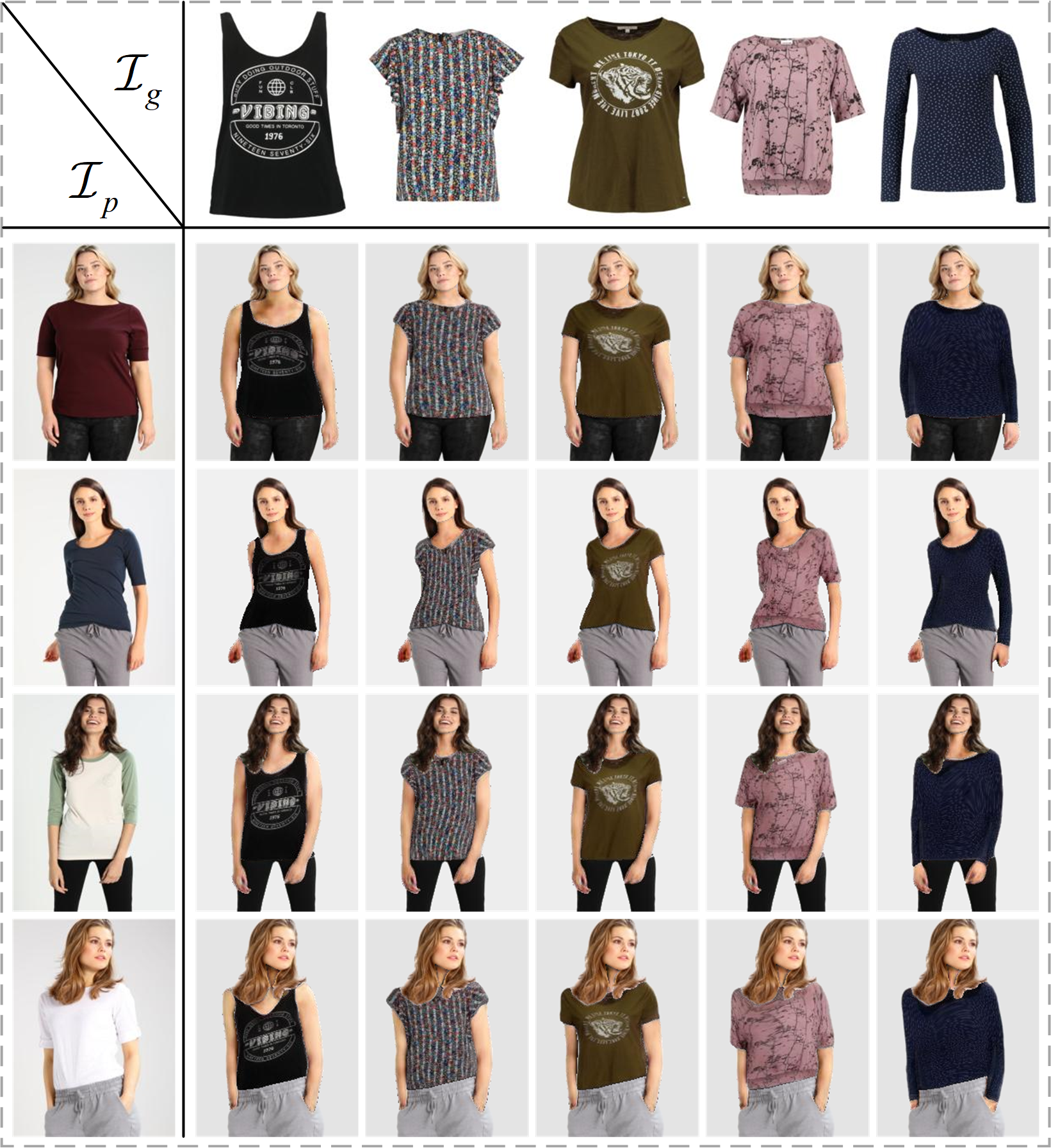}
\caption{The many-to-many virtual try-on.}
\label{fig17}
\end{figure}

Instead of trying a new garment on a single person, we show the instances of many-to-many virtual try-on. As shown in {\color{blue} Fig. \ref{fig17}}, we select five garments as ${{{\cal I}_g}}$ where their sleeves vary from short to long. And we select persons who wear a middle-sleeve garment as ${{{\cal I}_p}}$. By this means, our method shows its robustness under being with re-naked or covered skin.

To demonstrate the state-of-the-art performance of our method, we compare all baselines under two challenging virtual try-on scenarios: (1) Trying on a garment with different kinds of category; (2) The person has a complex pose where the arm blocks partial garment.

\begin{figure*}[!ht]
\centering
\includegraphics[scale=.24]{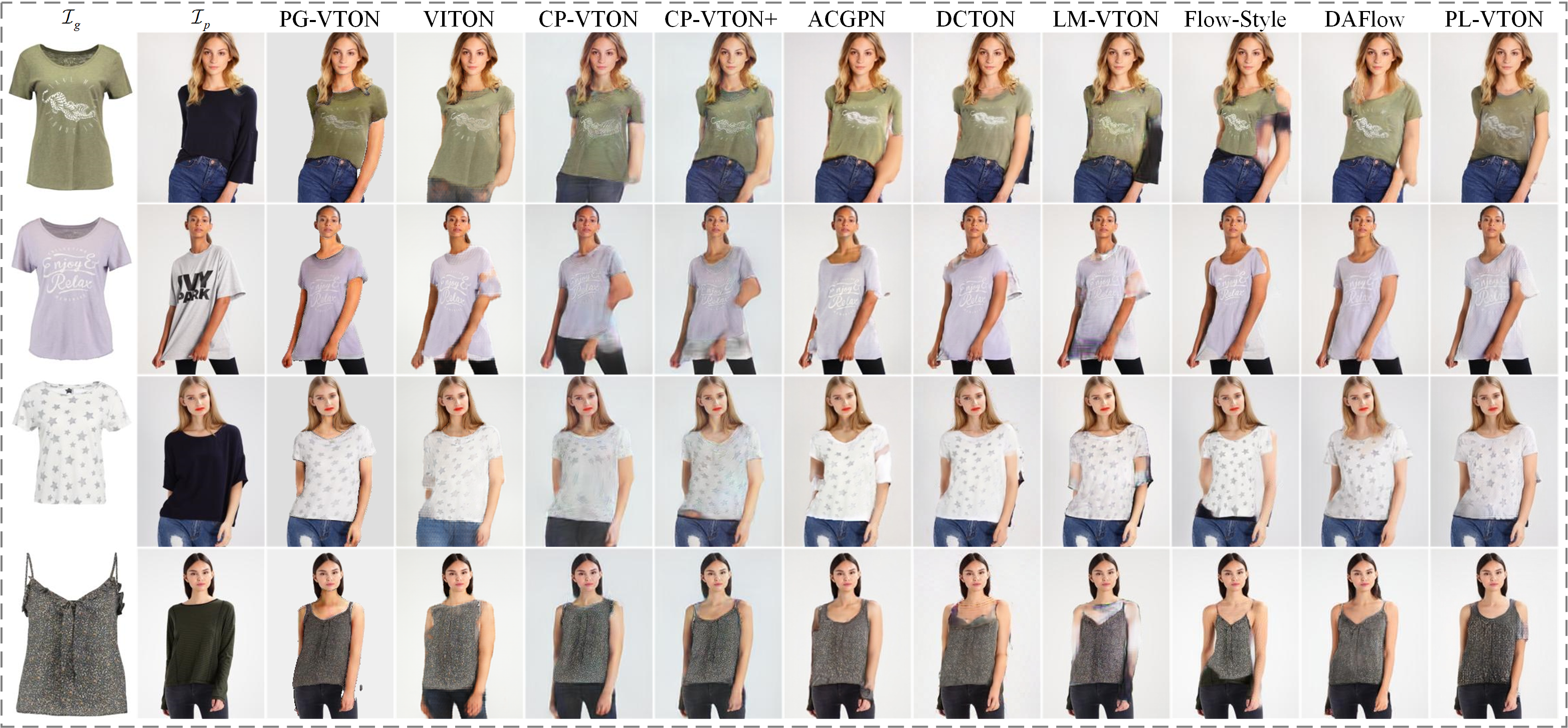}
\caption{The instances of trying on a garment with different kinds of category.}
\label{fig18}
\end{figure*}

\begin{figure*}[!ht]
\centering
\includegraphics[scale=.24]{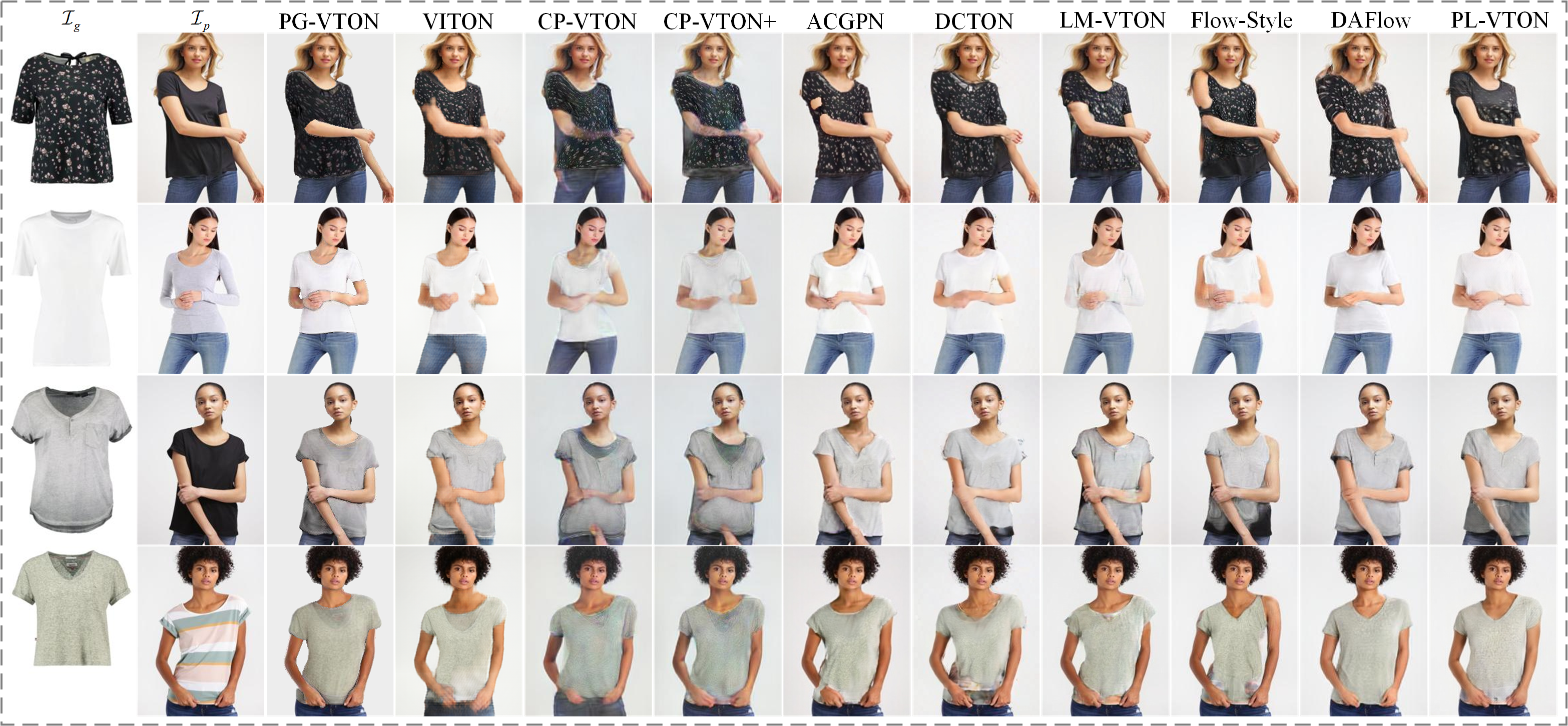}
\caption{The instances of being with a complex pose.}
\label{fig19}
\end{figure*}

As shown in {\color{blue} Fig. \ref{fig18}}, other baselines tend to rigidly alter the style of ${{{\cal I}_g}}$ as ${{{\cal I}_{pg}}}$ for the former virtual try-on circumstance. The infers unrealistic representations of the sleeve and arm, resulting in an inauthentic virtual try-on experience. In contrast, our method adopts the semantic category disentanglement in TPIM to alleviate the influence of the dressed garment, and adopts RSIM to address the re-naked skin inpainting issue. As a result, our method performs better in this scenario.

As shown in {\color{blue} Fig. \ref{fig19}}, other baselines fail to split the arm and garment from each other when they entangle together. Significant artifacts bring down the virtual try-on effect, and the texture reservation at the split garment part is also weakened. Conversely, our method is capable of inferring a reasonable try-on parsing even with a complex pose, and the fined mapping stage in PTM facilitates inferring the contents of split garment. Thus, our try-on result appears more authentic. In {\color{blue} Table \ref{table8}}, our method achieves 2.74 and 35.47 points for IS and hyperIQA. It demonstrates that our method has state-of-the-art performance in the overall image quality. The garment try-on and the color shift of the background category render significant variability in the similarity between ${{{\cal I}_t}}$ and ${{{\cal I}_g}}$ across various methods.

\begin{table*}[!ht]
\caption{The quantitative comparison between various methods.}
\label{table8}
\resizebox{2\columnwidth}{!}{
\centering
\begin{tabular}{ccccccccccc}
\toprule
                                                           &PG-VTON &VITON &CP-VTON &CP-VTON+ &ACGPN &DCTON & LM-VTON & Flow-Style & DAFlow & PL-VTON\\ \hline
        IS↑                                                & 2.74   &2.51  &2.60    &2.72     &2.61  &2.60  & 2.72    &2.64        &2.69    &2.54\\
       hyperIQA↑                                           &35.47   &32.81 &28.90   &28.72    &28.40 &30.54 &28.67    &29.18       &30.54   &29.90\\
       SSIM${\left( {{{\cal I}_t},{{\cal I}_g}} \right)}$↓ &0.785   &0.691 &0.694   &0.733    &0.767 &0.756 &0.814    &0.802       &0.737   &0.778\\
       FID${\left( {{{\cal I}_t},{{\cal I}_g}} \right)}$↓  &19.25   &39.85 &26.10   &22.55    &16.96 &16.49 &14.75    &16.26       &11.76   &14.40 \\\bottomrule
\end{tabular}
}
\end{table*}

We also invited 52 volunteers, comprising master and doctoral students, to undertake a subjective assessment of virtual try-on images. To ensure the quality and fairness of subjective assessment, each volunteer was presented with 6 group virtual try-on images, and they were instructed to select the image deemed best in terms of overall image quality, garment-wearing effect, and arm reserving. The sequence of methods within each group was randomly shuffled, and each volunteer dedicated more than 240 seconds to the subjective assessment. As shown in {\color{blue} Table \ref{table9}}, our method garnered a noteworthy 35.90\% majority preference for overall image quality, underscoring the efficacy of the progressive inference paradigm in enhancing the realism of try-on images. Our method also got 27.24\% and 22.44\% best votes for garment wearing effect and arm reserving, indicating that the designs of PTM and RSIM contribute significantly to the precise portrayal of local details in both garment and arm presentations.

\begin{table*}[!ht]
\caption{The vote rate in subjective assessment.}
\label{table9}
\resizebox{2\columnwidth}{!}{
\centering
\begin{tabular}{ccccccccccc}
\toprule
                                                           &PG-VTON  &VITON &CP-VTON &CP-VTON+ &ACGPN &DCTON & LM-VTON & Flow-Style & DAFlow & PL-VTON\\ \hline
        Overall Image Quality                              &35.90    &4.49  &1.28    &3.85     &3.21  &10.58 &2.56      &1.60       &16.03   &20.51\\
       Garment-Wearing Effect                              &27.24    &1.60  &2.88    &3.21     &4.49  &12.18 &10.58     &2.88       &22.76   &12.18\\
       Arm Reserving                                       &22.44    &2.88  &7.69    &2.88     &2.88  &6.09  &4.49      &6.09       &20.51   &24.04 \\\bottomrule
\end{tabular}
}
\end{table*}

 In nutshell, PG-VTON is more robust than other baselines for the universal try-on circumstances and preserves garment and skin details better. Moreover, we hope our method will contribute to further commercial applications.

\section{Conclusion}
\label{sec4}
This paper proposes a progressive virtual try-on method with a robust garment try-on strategy and specialized skin inpainting mechanism. To infer a reasonable try-on parsing, we disentangle the semantic categories in the person parsing to mitigate the influence of the original garment, and adopt U2Net as the backbone, which has proven to be more suitable for this particular task by comparison. By leveraging this shape guidance, we propose a garment try-on pipeline that adapts to the general case consisting of the coarse warping, the fined mapping, and composition. Specifically, we adopt the covering more and selecting one warping strategy and well-designed supervision strategy according to alignment. It enables the garment try-on to balance warping extent and texture reservation. To handle the re-naked skin inpainting in trying a new garment with a different category, we regulate StyleGAN2 with inputs of the target arm shape and spatial-agnostic intermediate code of skin. Moreover, we introduce random erasure into the self-supervised training and verify the influence of erasure extent on RSIM performance. Experiments demonstrate that our method has state-of-the-art performance under two challenging try-on scenarios when try-on a different-category garment or having a self-occlusion pose.

However, it is worth noting that PG-VTON is prior-based conditioned on pose and parsing. Though predicted models of human pose and parsing have the state-of-the art performance, it is inevitable to introduce additional errors into our method. Conversely, the prediction precision increase of human parsing will contribute to the performance of our method, especially for TPIM. In further work, we will devote ourselves to enabling our method to be prior-free without changing its robust performance and promote the further commercial application of image-based virtual try-on.

\footnotesize
\bibliographystyle{IEEEtran}
\bibliography{refs}{}

\begin{thebibliography}{10}
\providecommand{\url}[1]{#1}
\csname url@samestyle\endcsname
\providecommand{\newblock}{\relax}
\providecommand{\bibinfo}[2]{#2}
\providecommand{\BIBentrySTDinterwordspacing}{\spaceskip=0pt\relax}
\providecommand{\BIBentryALTinterwordstretchfactor}{4}
\providecommand{\BIBentryALTinterwordspacing}{\spaceskip=\fontdimen2\font plus
\BIBentryALTinterwordstretchfactor\fontdimen3\font minus
  \fontdimen4\font\relax}
\providecommand{\BIBforeignlanguage}[2]{{%
\expandafter\ifx\csname l@#1\endcsname\relax
\typeout{** WARNING: IEEEtran.bst: No hyphenation pattern has been}%
\typeout{** loaded for the language `#1'. Using the pattern for}%
\typeout{** the default language instead.}%
\else
\language=\csname l@#1\endcsname
\fi
#2}}
\providecommand{\BIBdecl}{\relax}
\BIBdecl

\bibitem{hu20213dbodynet}
P.~Hu, E.~S.-L. Ho, and A.~Munteanu, ``3dbodynet: fast reconstruction of 3d
  animatable human body shape from a single commodity depth camera,''
  \emph{IEEE Transactions on Multimedia}, vol.~24, pp. 2139--2149, 2021.

\bibitem{zhao20183}
T.~Zhao, S.~Li, K.~N. Ngan, and F.~Wu, ``3-d reconstruction of human body shape
  from a single commodity depth camera,'' \emph{IEEE Transactions on
  Multimedia}, vol.~21, no.~1, pp. 114--123, 2018.

\bibitem{sekhavat2016privacy}
Y.~A. Sekhavat, ``Privacy preserving cloth try-on using mobile augmented
  reality,'' \emph{IEEE Transactions on Multimedia}, vol.~19, no.~5, pp.
  1041--1049, 2016.

\bibitem{han2018viton}
X.~Han, Z.~Wu, Z.~Wu, R.~Yu, and L.~S. Davis, ``Viton: An image-based virtual
  try-on network,'' in \emph{Proceedings of the IEEE/CVF Conference on Computer
  Vision and Pattern Recognition(CVPR)}, 2018, pp. 7543--7552.

\bibitem{wang2018toward}
B.~Wang, H.~Zheng, X.~Liang, Y.~Chen, L.~Lin, and M.~Yang, ``Toward
  characteristic-preserving image-based virtual try-on network,'' in
  \emph{Proceedings of the European Conference on Computer Vision (ECCV)},
  2018, pp. 589--604.

\bibitem{yu2019vtnfp}
R.~Yu, X.~Wang, and X.~Xie, ``Vtnfp: An image-based virtual try-on network with
  body and clothing feature preservation,'' in \emph{Proceedings of the
  IEEE/CVF International Conference on Computer Vision (ICCV)}, 2019, pp.
  10\,511--10\,520.

\bibitem{yang2020towards}
H.~Yang, R.~Zhang, X.~Guo, W.~Liu, W.~Zuo, and P.~Luo, ``Towards
  photo-realistic virtual try-on by adaptively generating-preserving image
  content,'' in \emph{Proceedings of the IEEE/CVF Conference on Computer Vision
  and Pattern Recognition (CVPR)}, 2020, pp. 7850--7859.

\bibitem{minar2020cp}
M.~R. Minar, T.~T. Tuan, H.~Ahn, P.~Rosin, and Y.-K. Lai, ``Cp-vton+: Clothing
  shape and texture preserving image-based virtual try-on,'' in \emph{CVPR
  Workshops}, 2020.

\bibitem{neuberger2020image}
A.~Neuberger, E.~Borenstein, B.~Hilleli, E.~Oks, and S.~Alpert, ``Image based
  virtual try-on network from unpaired data,'' in \emph{Proceedings of the
  IEEE/CVF Conference on Computer Vision and Pattern Recognition (CVPR)}, 2020,
  pp. 5184--5193.

\bibitem{ge2021disentangled}
C.~Ge, Y.~Song, Y.~Ge, H.~Yang, W.~Liu, and P.~Luo, ``Disentangled cycle
  consistency for highly-realistic virtual try-on,'' in \emph{Proceedings of
  the IEEE/CVF Conference on Computer Vision and Pattern Recognition (CVPR)},
  2021, pp. 16\,928--16\,937.

\bibitem{he2022style}
S.~He, Y.-Z. Song, and T.~Xiang, ``Style-based global appearance flow for
  virtual try-on,'' in \emph{Proceedings of the IEEE/CVF Conference on Computer
  Vision and Pattern Recognition (CVPR)}, 2022, pp. 3470--3479.

\bibitem{bai2022single}
S.~Bai, H.~Zhou, Z.~Li, C.~Zhou, and H.~Yang, ``Single stage virtual try-on via
  deformable attention flows,'' in \emph{Proceedings of the European Conference
  on Computer Vision (ECCV)}.\hskip 1em plus 0.5em minus 0.4em\relax Springer,
  2022, pp. 409--425.

\bibitem{fang2022novel}
N.~Fang, L.~Qiu, S.~Zhang, Z.~Wang, K.~Hu, and L.~Dong, ``A novel human image
  sequence synthesis method by pose-shape-content inference,'' \emph{IEEE
  Transactions on Multimedia}, vol.~25, pp. 6512--6524, 2023.

\bibitem{ma2021fda}
L.~Ma, K.~Huang, D.~Wei, Z.-Y. Ming, and H.~Shen, ``Fda-gan: Flow-based dual
  attention gan for human pose transfer,'' \emph{IEEE Transactions on
  Multimedia}, 2021.

\bibitem{hu2022spg}
B.~Hu, P.~Liu, Z.~Zheng, and M.~Ren, ``Spg-vton: Semantic prediction guidance
  for multi-pose virtual try-on,'' \emph{IEEE Transactions on Multimedia},
  vol.~24, pp. 1233--1246, 2022.

\bibitem{dong2019towards}
H.~Dong, X.~Liang, X.~Shen, B.~Wang, H.~Lai, J.~Zhu, Z.~Hu, and J.~Yin,
  ``Towards multi-pose guided virtual try-on network,'' in \emph{Proceedings of
  the IEEE/CVF International Conference on Computer Vision (ICCV)}, 2019, pp.
  9026--9035.

\bibitem{xie2021towards}
Z.~Xie, Z.~Huang, F.~Zhao, H.~Dong, M.~Kampffmeyer, and X.~Liang, ``Towards
  scalable unpaired virtual try-on via patch-routed spatially-adaptive gan,''
  in \emph{Advances in Neural Information Processing Systems (NeurIPS)},
  vol.~34, 2021, pp. 2598--2610.

\bibitem{cui2021dressing}
A.~Cui, D.~McKee, and S.~Lazebnik, ``Dressing in order: Recurrent person image
  generation for pose transfer, virtual try-on and outfit editing,'' in
  \emph{Proceedings of the IEEE/CVF International Conference on Computer Vision
  (ICCV)}, 2021, pp. 14\,638--14\,647.

\bibitem{raj2018swapnet}
A.~Raj, P.~Sangkloy, H.~Chang, J.~Hays, D.~Ceylan, and J.~Lu, ``Swapnet: Image
  based garment transfer,'' in \emph{Proceedings of the European Conference on
  Computer Vision (ECCV)}.\hskip 1em plus 0.5em minus 0.4em\relax Springer,
  2018, pp. 679--695.

\bibitem{liu2019swapgan}
Y.~Liu, W.~Chen, L.~Liu, and M.~S. Lew, ``Swapgan: A multistage generative
  approach for person-to-person fashion style transfer,'' \emph{IEEE
  Transactions on Multimedia}, vol.~21, no.~9, pp. 2209--2222, 2019.

\bibitem{liu2021spatial}
T.~Liu, J.~Zhang, X.~Nie, Y.~Wei, S.~Wei, Y.~Zhao, and J.~Feng, ``Spatial-aware
  texture transformer for high-fidelity garment transfer,'' \emph{IEEE
  Transactions on Image Processing}, vol.~30, pp. 7499--7510, 2021.

\bibitem{lewis2021tryongan}
K.~M. Lewis, S.~Varadharajan, and I.~Kemelmacher-Shlizerman, ``Tryongan:
  Body-aware try-on via layered interpolation,'' \emph{ACM Transactions on
  Graphics (TOG)}, vol.~40, no.~4, pp. 1--10, 2021.

\bibitem{dong2019fw}
H.~Dong, X.~Liang, X.~Shen, B.~Wu, B.-C. Chen, and J.~Yin, ``Fw-gan:
  Flow-navigated warping gan for video virtual try-on,'' in \emph{Proceedings
  of the IEEE/CVF International Conference on Computer Vision (ICCV)}, 2019,
  pp. 1161--1170.

\bibitem{chen2021fashionmirror}
C.-Y. Chen, L.~Lo, P.-J. Huang, H.-H. Shuai, and W.-H. Cheng, ``Fashionmirror:
  Co-attention feature-remapping virtual try-on with sequential template
  poses,'' in \emph{Proceedings of the IEEE/CVF International Conference on
  Computer Vision (ICCV)}, 2021, pp. 13\,809--13\,818.

\bibitem{choi2021viton}
S.~Choi, S.~Park, M.~Lee, and J.~Choo, ``Viton-hd: High-resolution virtual
  try-on via misalignment-aware normalization,'' in \emph{Proceedings of the
  IEEE/CVF Conference on Computer Vision and Pattern Recognition (CVPR)}, 2021,
  pp. 14\,131--14\,140.

\bibitem{issenhuth2020not}
T.~Issenhuth, J.~Mary, and C.~Calauz{\`e}nes, ``Do not mask what you do not
  need to mask: a parser-free virtual try-on,'' in \emph{Proceedings of the
  European Conference on Computer Vision (ECCV)}.\hskip 1em plus 0.5em minus
  0.4em\relax Springer, 2020, pp. 619--635.

\bibitem{ge2021parser}
Y.~Ge, Y.~Song, R.~Zhang, C.~Ge, W.~Liu, and P.~Luo, ``Parser-free virtual
  try-on via distilling appearance flows,'' in \emph{Proceedings of the
  IEEE/CVF Conference on Computer Vision and Pattern Recognition (CVPR)}, 2021,
  pp. 8485--8493.

\bibitem{jaderberg2015advances}
M.~Jaderberg, K.~Simonyan, A.~Zisserman, and k.~kavukcuoglu, ``Spatial
  transformer networks,'' in \emph{Advances in Neural Information Processing
  Systems (NeurIPS)}, vol.~28.\hskip 1em plus 0.5em minus 0.4em\relax Curran
  Associates, Inc., 2015.

\bibitem{bookstein1989principal}
F.~L. Bookstein, ``Principal warps: Thin-plate splines and the decomposition of
  deformations,'' \emph{IEEE Transactions on Pattern Analysis and Machine
  Intelligence}, vol.~11, no.~6, pp. 567--585, 1989.

\bibitem{dosovitskiy2015flownet}
A.~Dosovitskiy, P.~Fischer, E.~Ilg, P.~Hausser, C.~Hazirbas, V.~Golkov, P.~Van
  Der~Smagt, D.~Cremers, and T.~Brox, ``Flownet: Learning optical flow with
  convolutional networks,'' in \emph{Proceedings of the IEEE/CVF International
  Conference on Computer Vision (ICCV)}, 2015, pp. 2758--2766.

\bibitem{ronneberger2015u}
O.~Ronneberger, P.~Fischer, and T.~Brox, ``U-net: Convolutional networks for
  biomedical image segmentation,'' in \emph{International Conference on Medical
  image computing and computer-assisted intervention}.\hskip 1em plus 0.5em
  minus 0.4em\relax Springer, 2015, pp. 234--241.

\bibitem{dosovitskiy2020image}
A.~Dosovitskiy, L.~Beyer, A.~Kolesnikov, D.~Weissenborn, X.~Zhai,
  T.~Unterthiner, M.~Dehghani, M.~Minderer, G.~Heigold, S.~Gelly \emph{et~al.},
  ``An image is worth 16x16 words: Transformers for image recognition at
  scale,'' in \emph{International Conference on Learning Representations
  (ICLR)}, 2020.

\bibitem{liu2021swin}
Z.~Liu, Y.~Lin, Y.~Cao, H.~Hu, Y.~Wei, Z.~Zhang, S.~Lin, and B.~Guo, ``Swin
  transformer: Hierarchical vision transformer using shifted windows,'' in
  \emph{Proceedings of the IEEE/CVF International Conference on Computer Vision
  (ICCV)}, 2021, pp. 10\,012--10\,022.

\bibitem{yang2021focal}
J.~Yang, C.~Li, P.~Zhang, X.~Dai, B.~Xiao, L.~Yuan, and J.~Gao, ``Focal
  attention for long-range interactions in vision transformers,'' in
  \emph{Advances in Neural Information Processing Systems (NeurIPS)},
  vol.~34.\hskip 1em plus 0.5em minus 0.4em\relax Curran Associates, Inc.,
  2021, pp. 30\,008--30\,022.

\bibitem{zamir2022restormer}
S.~W. Zamir, A.~Arora, S.~Khan, M.~Hayat, F.~S. Khan, and M.-H. Yang,
  ``Restormer: Efficient transformer for high-resolution image restoration,''
  in \emph{Proceedings of the IEEE/CVF Conference on Computer Vision and
  Pattern Recognition (CVPR)}, 2022, pp. 5728--5739.

\bibitem{karras2020analyzing}
T.~Karras, S.~Laine, M.~Aittala, J.~Hellsten, J.~Lehtinen, and T.~Aila,
  ``Analyzing and improving the image quality of stylegan,'' in
  \emph{Proceedings of the IEEE/CVF Conference on Computer Vision and Pattern
  Recognition (CVPR)}, 2020, pp. 8110--8119.

\bibitem{guler2018densepose}
R.~A. G{\"u}ler, N.~Neverova, and I.~Kokkinos, ``Densepose: Dense human pose
  estimation in the wild,'' in \emph{Proceedings of the IEEE/CVF Conference on
  Computer Vision and Pattern Recognition (CVPR)}, 2018, pp. 7297--7306.

\bibitem{he2020grapy}
H.~He, J.~Zhang, Q.~Zhang, and D.~Tao, ``Grapy-ml: Graph pyramid mutual
  learning for cross-dataset human parsing,'' in \emph{Proceedings of the AAAI
  Conference on Artificial Intelligence (AAAI)}, vol.~34, no.~07, 2020, pp.
  10\,949--10\,956.

\bibitem{qin2020u2}
X.~Qin, Z.~Zhang, C.~Huang, M.~Dehghan, O.~R. Zaiane, and M.~Jagersand,
  ``U2-net: Going deeper with nested u-structure for salient object
  detection,'' \emph{Pattern Recognition}, vol. 106, p. 107404, 2020.

\bibitem{simonyan2014very}
K.~Simonyan and A.~Zisserman, ``Very deep convolutional networks for
  large-scale image recognition,'' \emph{arXiv preprint arXiv:1409.1556}, 2014.

\bibitem{peebles2022gan}
W.~Peebles, J.-Y. Zhu, R.~Zhang, A.~Torralba, A.~A. Efros, and E.~Shechtman,
  ``Gan-supervised dense visual alignment,'' in \emph{Proceedings of the
  IEEE/CVF Conference on Computer Vision and Pattern Recognition (CVPR)}, 2022,
  pp. 13\,470--13\,481.

\bibitem{zhong2020random}
Z.~Zhong, L.~Zheng, G.~Kang, S.~Li, and Y.~Yang, ``Random erasing data
  augmentation,'' in \emph{Proceedings of the AAAI Conference on Artificial
  Intelligence (AAAI)}, vol.~34, no.~07, 2020, pp. 13\,001--13\,008.

\bibitem{alom2019recurrent}
M.~Z. Alom, C.~Yakopcic, M.~Hasan, T.~M. Taha, and V.~K. Asari, ``Recurrent
  residual u-net for medical image segmentation,'' \emph{Journal of Medical
  Imaging}, vol.~6, no.~1, p. 014006, 2019.

\bibitem{zhang2018road}
Z.~Zhang, Q.~Liu, and Y.~Wang, ``Road extraction by deep residual u-net,''
  \emph{IEEE Geoscience and Remote Sensing Letters}, vol.~15, no.~5, pp.
  749--753, 2018.

\bibitem{zhou2019unet++}
Z.~Zhou, M.~M.~R. Siddiquee, N.~Tajbakhsh, and J.~Liang, ``Unet++: Redesigning
  skip connections to exploit multiscale features in image segmentation,''
  \emph{IEEE Transactions on Medical Imaging}, vol.~39, no.~6, pp. 1856--1867,
  2019.

\bibitem{oktay2018attention}
O.~Oktay, J.~Schlemper, L.~L. Folgoc, M.~Lee, M.~Heinrich, K.~Misawa, K.~Mori,
  S.~McDonagh, N.~Y. Hammerla, B.~Kainz \emph{et~al.}, ``Attention u-net:
  Learning where to look for the pancreas,'' \emph{arXiv preprint
  arXiv:1804.03999}, 2018.

\bibitem{zheng2021rethinking}
S.~Zheng, J.~Lu, H.~Zhao, X.~Zhu, Z.~Luo, Y.~Wang, Y.~Fu, J.~Feng, T.~Xiang,
  P.~H. Torr \emph{et~al.}, ``Rethinking semantic segmentation from a
  sequence-to-sequence perspective with transformers,'' in \emph{Proceedings of
  the IEEE/CVF Conference on Computer Vision and Pattern Recognition (CVPR)},
  2021, pp. 6881--6890.

\bibitem{cao2021swin}
H.~Cao, Y.~Wang, J.~Chen, D.~Jiang, X.~Zhang, Q.~Tian, and M.~Wang,
  ``Swin-unet: Unet-like pure transformer for medical image segmentation,''
  \emph{arXiv preprint arXiv:2105.05537}, 2021.

\bibitem{cao2017realtime}
Z.~Cao, T.~Simon, S.-E. Wei, and Y.~Sheikh, ``Realtime multi-person 2d pose
  estimation using part affinity fields,'' in \emph{Proceedings of the IEEE/CVF
  Conference on Computer Vision and Pattern Recognition (CVPR)}, 2017, pp.
  7291--7299.

\bibitem{salimans2016improved}
T.~Salimans, I.~Goodfellow, W.~Zaremba, V.~Cheung, A.~Radford, and X.~Chen,
  ``Improved techniques for training gans,'' in \emph{Advances in Neural
  Information Processing Systems (NeurIPS)}, 2016, pp. 2234--2242.

\bibitem{karras2019style}
T.~Karras, S.~Laine, and T.~Aila, ``A style-based generator architecture for
  generative adversarial networks,'' in \emph{Proceedings of the IEEE/CVF
  Conference on Computer Vision and Pattern Recognition (CVPR)}, 2019, pp.
  4401--4410.

\bibitem{liu2021toward}
G.~Liu, D.~Song, R.~Tong, and M.~Tang, ``Toward realistic virtual try-on
  through landmark guided shape matching,'' in \emph{Proceedings of the AAAI
  Conference on Artificial Intelligence (AAAI)}, vol.~35, no.~3, 2021, pp.
  2118--2126.

\bibitem{han2022progressive}
X.~Han, S.~Zhang, Q.~Liu, Z.~Li, and C.~Wang, ``Progressive limb-aware virtual
  try-on,'' in \emph{Proceedings of the 30th ACM International Conference on
  Multimedia}, 2022, pp. 2420--2429.

\bibitem{su2020blindly}
S.~Su, Q.~Yan, Y.~Zhu, C.~Zhang, X.~Ge, J.~Sun, and Y.~Zhang, ``Blindly assess
  image quality in the wild guided by a self-adaptive hyper network,'' in
  \emph{Proceedings of the IEEE/CVF Conference on Computer Vision and Pattern
  Recognition (CVPR)}, 2020, pp. 3667--3676.

\end{thebibliography}

\vspace{-28 mm}
\begin{IEEEbiography}[{\includegraphics[width=1in,height=1.25in,clip,keepaspectratio]{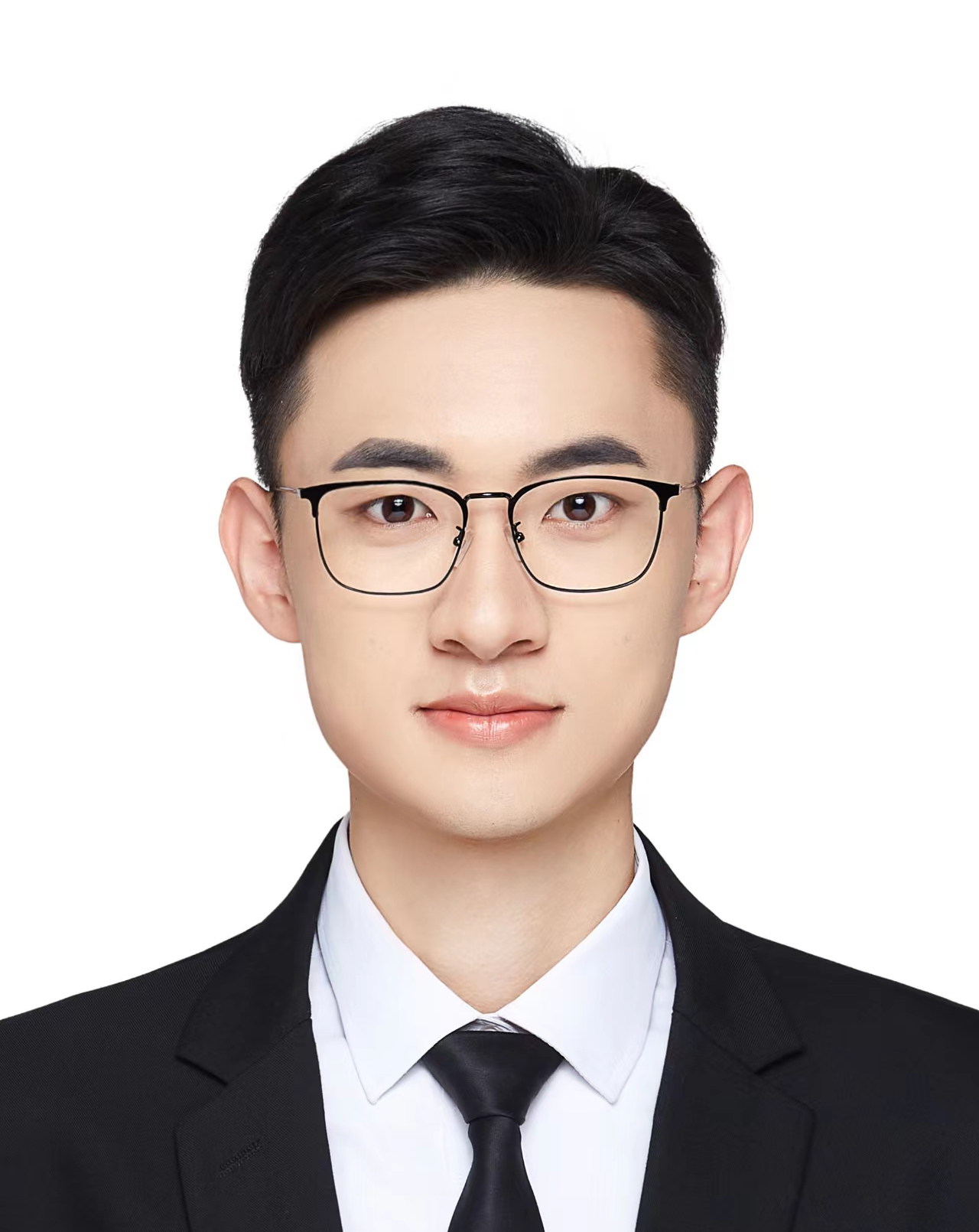}}]{Naiyu Fang}
(Graduate Student Member, IEEE) received the B.Eng. degree in the School of Mechanical Engineering from Dalian University of Technology, Dalian, China, in 2019. He is currently a Ph.D. candidate in the School of Mechanical Engineering, Zhejiang University, China. His research interests include virtual try-on and autonomous driving.\end{IEEEbiography}
\vspace{-25 mm}

\begin{IEEEbiography}[{\includegraphics[width=1in,height=1.25in,clip,keepaspectratio]{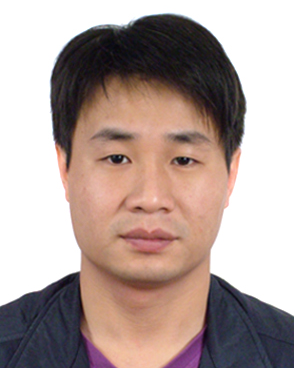}}]{Lemiao Qiu}
(Member, IEEE) received the Ph.D. degree from the Department of Mechanical Engineering, Zhejiang University, Hangzhou, China, in 2008. He is currently an Professor at the Department of Mechanical Engineering, Zhejiang University, China. His research interests include computer graphics, virtual try-on, and production informatization.\end{IEEEbiography}

\vspace{-25 mm}
\begin{IEEEbiography}[{\includegraphics[width=1in,height=1.25in,clip,keepaspectratio]{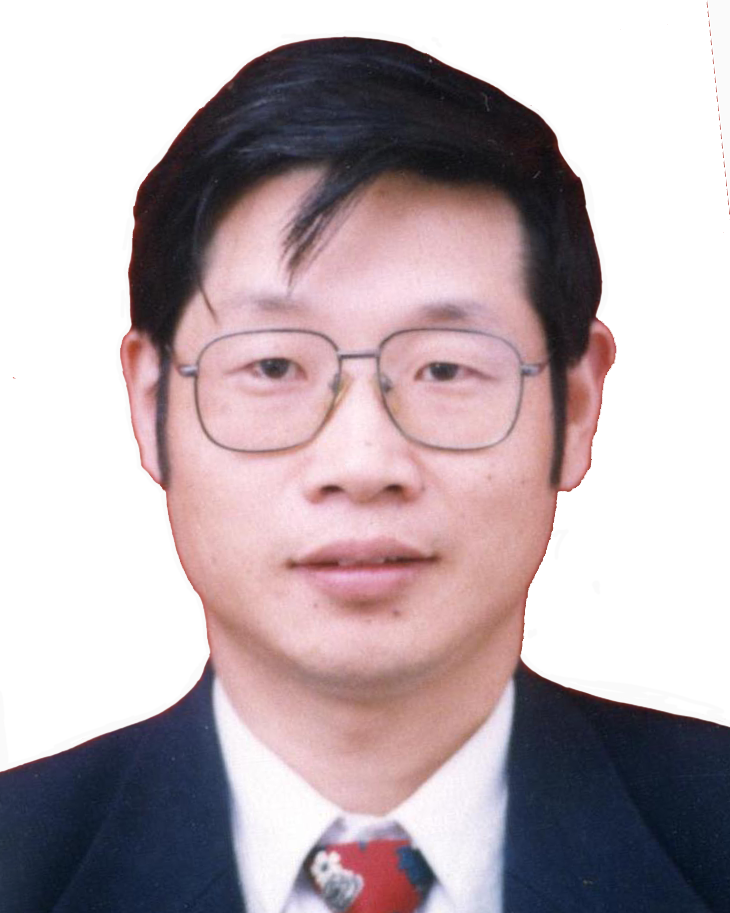}}]{Shuyou Zhang}
is currently a Distinguished Professor and a Ph.D. Supervisor at the Department of Mechanical Engineering, Zhejiang University, China. His research interests include computer graphics, computer vision, and product digital design.\end{IEEEbiography}
\vspace{-25 mm}

\begin{IEEEbiography}[{\includegraphics[width=1in,height=1.25in,clip,keepaspectratio]{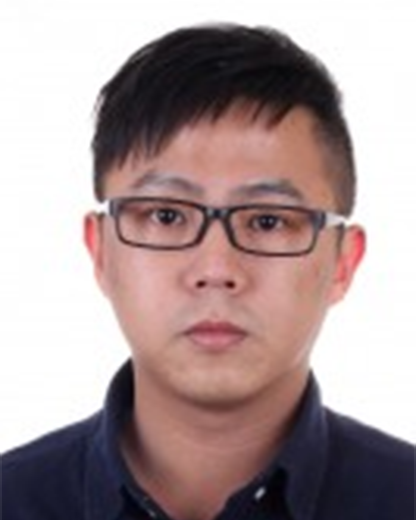}}]{Zili Wang}
(Member, IEEE) received the Ph.D. degree from the Department of Mechanical Engineering, Zhejiang University, Hangzhou, China in 2018. He is currently Research Associate at the Department of Mechanical Engineering, Zhejiang University, Hangzhou, China. His research interests include computer-aided design and computer graphics.\end{IEEEbiography}

\vspace{-25 mm}

\begin{IEEEbiography}[{\includegraphics[width=1in,height=1.25in,clip,keepaspectratio]{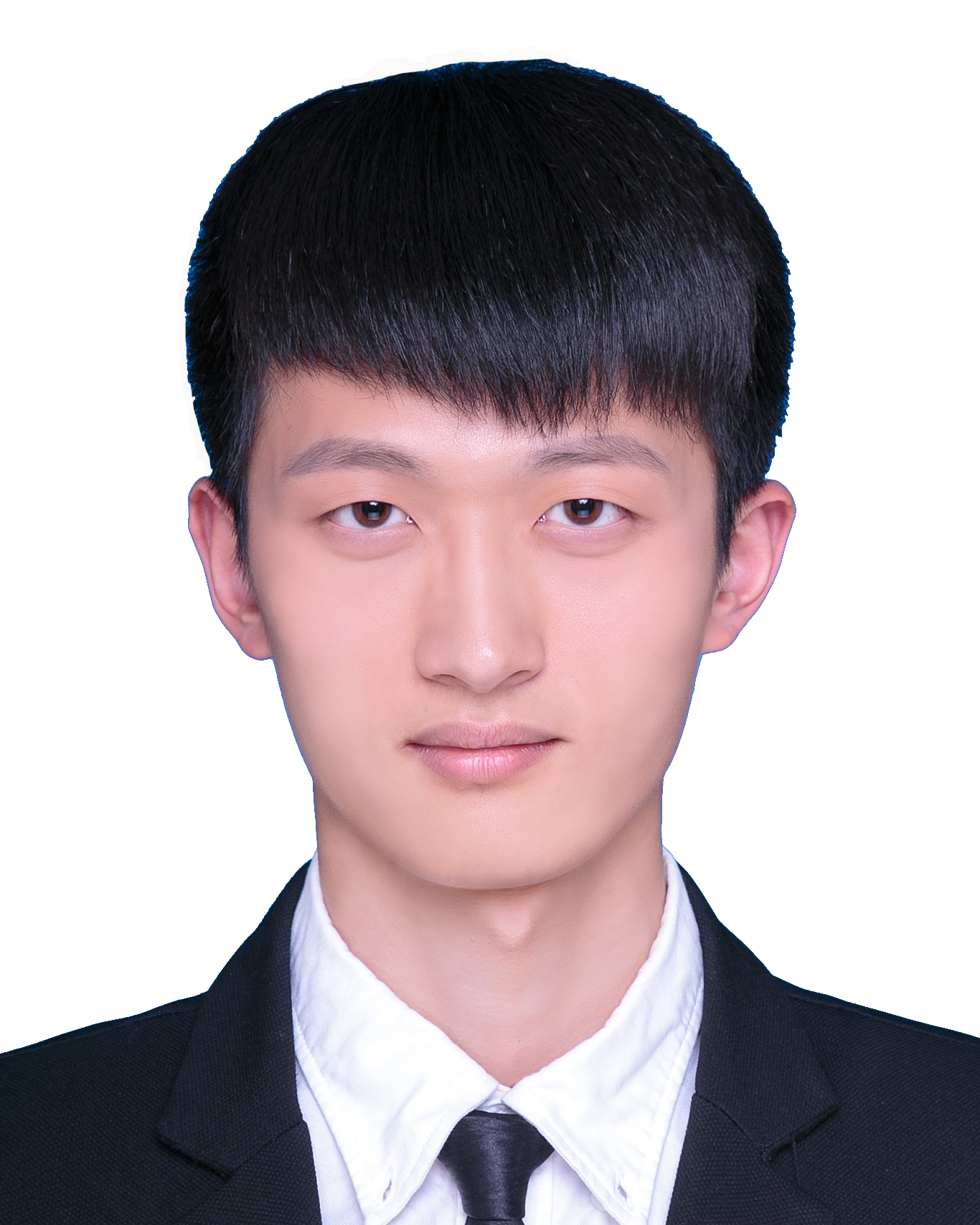}}]{Kerui Hu}
received the B.Eng. degree in the School of China University of Mining and Technology, Xuzhou, China, in 2018. He is currently a Ph.D. candidate in the School of Mechanical Engineering, Zhejiang University, China. His research interests include data mining and collaborative filtering.

\vspace{-25 mm}
\end{IEEEbiography}

\end{document}